\documentclass[screen,acmsmall]{acmart}

\usepackage{amsmath,amsfonts,bm}

\def\eqref#1{equation~\ref{#1}}

\def\1{\bm{1}}

\def\rmH{{\mathbf{H}}}

\def\vh{{\bm{h}}}

\def\vp{{\bm{p}}}

\DeclareMathAlphabet{\mathsfit}{\encodingdefault}{\sfdefault}{m}{sl}
\SetMathAlphabet{\mathsfit}{bold}{\encodingdefault}{\sfdefault}{bx}{n}

\def\gD{{\mathcal{D}}}
\def\gE{{\mathcal{E}}}

\def\gG{{\mathcal{G}}}

\def\gL{{\mathcal{L}}}

\def\gV{{\mathcal{V}}}

\def\gX{{\mathcal{X}}}

\newcommand{\softmax}{\mathrm{softmax}}

\usepackage{url}
\usepackage{graphicx}
\usepackage{color}
\usepackage{multirow}
\usepackage{subfigure}
\usepackage{ulem}
\usepackage{enumitem}
\usepackage{makecell}
\usepackage{tabularx}
\usepackage{threeparttable}
\usepackage{arydshln}
\usepackage{soul}
\usepackage{color, xcolor}
\usepackage[utf8]{inputenc}
\usepackage{tcolorbox}
\definecolor{mylightgreen}{RGB}{200,255,200}
\usepackage{marvosym}
\usepackage{microtype}
\usepackage{graphicx}
\usepackage{adjustbox}
\usepackage{array}

\usepackage{algorithm}
\usepackage{algpseudocode}

\DeclareMathOperator*{\mlp}{MLP}
\DeclareMathOperator*{\transformerenc}{Transformer-Enc}
\DeclareMathOperator*{\pool}{Pool}

\AtBeginDocument{%
  }

\setcopyright{acmlicensed}
\copyrightyear{2026}
\acmYear{2026}
\acmDOI{XXXXXXX.XXXXXXX}

\acmBooktitle{ACM Trans. Inf. Syst.}
\acmISBN{978-1-4503-XXXX-X/18/06}

\begin{document}

\title{
A Graph-Enhanced Defense Framework for Explainable Fake News Detection with LLM}


\author{Bo Wang}
\orcid{https://orcid.org/0000-0001-7158-7046}
\affiliation{%
  \institution{School of Artificial Intelligence, Jilin University}
  \city{Changchun}
  \country{China}}

\email{wangbo21@mails.jlu.edu}

\author{Jing Ma}
\authornote{Corresponding author.}
\orcid{https://orcid.org/0000-0002-7464-8331}
\author{Hongzhan Lin}
\orcid{https://orcid.org/0000-0002-4111-8334}
\affiliation{%
  \institution{Hong Kong Baptist University}
  \city{Hong Kong SAR}
  \country{China}
  }
\email{majing@comp.hkbu.edu.hk}
\email{cshzlin@comp.hkbu.edu.hk}

\author{Zhiwei Yang}
\orcid{https://orcid.org/0000-0002-0534-158X}
\affiliation{%
  \institution{Guangdong Institute of Smart Education, Jinan University}
  \city{Guangzhou}
  \country{China}
  }
\email{yangzw18@mails.jlu.edu.cn}

\author{Ruichao Yang}
\orcid{https://orcid.org/0000-0003-3749-3622}
\affiliation{%
  \institution{University of Science and Technology Beijing}
  \city{Beijing}
  \country{China}
  }
\email{yangruichao@ustb.edu.cn}

\author{Yuan Tian}
\authornotemark[1]
\orcid{https://orcid.org/0000-0003-1242-6714}
\affiliation{%
  \institution{School of Artificial Intelligence, Jilin University}
  \city{Changchun}
  \country{China;}
  \institution{Engineering Research Center of Knowledge-Driven Human-Machine Intelligence}
  \city{MoE}
  \country{China}
  }
\email{yuantian@jlu.edu.cn}

\author{Yi Chang}
\authornotemark[1]
\orcid{https://orcid.org/0000-0003-2697-8093}
\affiliation{%
  \institution{School of Artificial Intelligence, Jilin University}
  \city{Changchun}
  \country{China;}
  \institution{Engineering Research Center of Knowledge-Driven Human-Machine Intelligence}
  \city{MoE}
  \country{China;}
  \institution{International Center of Future Science, Jilin University}
  \city{Changchun}
  \country{China}
  }
\email{yichang@jlu.edu.cn}


\renewcommand{\shortauthors}{B.Wang and J.Ma et al.}

\begin{abstract}

Explainable fake news detection aims to assess the veracity of news claims while providing human-friendly explanations. Existing methods incorporating investigative journalism are often inefficient and struggle with breaking news. Recent advances in large language models (LLMs) enable leveraging externally retrieved reports as evidence for detection and explanation generation, but unverified reports may introduce inaccuracies. Moreover, effective explainable fake news detection should provide a comprehensible explanation for all aspects of a claim to assist the public in verifying its accuracy. To address these challenges, we propose a graph-enhanced defense framework (G-Defense)  that provides fine-grained explanations based solely on unverified reports. Specifically, we construct a claim-centered graph by decomposing the news claim into several sub-claims and modeling their dependency relationships. For each sub-claim, we use the retrieval-augmented generation (RAG) technique to retrieve salient evidence and generate competing explanations. We then introduce a defense-like inference module based on the graph to assess the overall veracity. Finally, we prompt an LLM to generate an intuitive explanation graph. Experimental results demonstrate that G-Defense achieves state-of-the-art performance in both veracity detection and the quality of its explanations.
\end{abstract}




\maketitle

\section{Introduction}

The rapid dissemination of fake news on social media platforms has become a critical issue, posing severe risks to both individuals and society. 
For instance, during the early stage of the COVID-19 pandemic, a widely circulated fake news that ``\textit{Face masks drastically reduce oxygen intake and cause carbon dioxide toxicity.}\footnote{\url{https://www.bbc.com/news/53108405}}'' led many people to refuse to wear masks, which made the virus spread faster and put extra pressure on healthcare systems around the world. Later, trusted scientific organizations disproved this claim and provided detailed justifications, helping to rebuild public trust in mask usage. This case suggests how important solid justification is in reducing the social harm caused by fake news. 
However, merely relying on traditional journalism to detect fake news is not practical, since it is labor-intensive and time-consuming, which results in limited coverage and delays in debunking. Thus, it demands an automated method, aiming not only to detect fake news but also to provide timely and clear justifications.

Many studies have explored explainable fake news detection, which aims to perform veracity detection and generate explanations simultaneously \cite{EFKD_survey_KotonyaT_coling20}. 
Initially, some works use the relevant reports as supporting data, and then highlight discrete words, phrases, or sentences from them that have high contributions to the final prediction, to bring some content of interpretability \cite{DeClarE_EMNLP18, conflict_wu_aaai21,Nie_AAAI19,dEFEND_KDD19}. However, these fragmentary explanations lack the intuitiveness\footnote{Intuitiveness reflects how easily a human can understand and follow the explanation.
} and credibility\footnote{Credibility denotes the perceived trustworthiness or reliability of the information \cite{castillo2011information}.} needed for clear understanding. 
Motivated by the effectiveness of human justifications in veracity prediction \cite{LIAR-PLUS_Alhindi_18}, \citet{GenFE_ACL20} proposed to generate explanations by summarizing debunked reports. While this method improves explainability, it inevitably suffers from delayed detection and explanation generation, as debunked reports are not always available and typically require substantial human effort, which also makes the approach inefficient. 
More recently, inspired by the effectiveness of the wisdom of crowds in fact-checking \cite{crowds_wisdom-Allen_2021}, \citet{zhiwei22coling} proposed leveraging externally retrieved raw reports containing collective wisdom as candidate evidence for both detection and explanation. 
While these reports providing diverse perspectives can support veracity prediction, the generated explanations are still produced in an extraction-based manner. 
As a result, the selected pieces of evidence are often discrete, lacking a coherent reasoning process and failing to produce explanations that are fully aligned with the model's final decision.

Large language models (LLMs) offer a promising direction to this issue, as they have demonstrated strong capabilities in language processing  \cite{InstructGPT_ouyang_NIPS22,GPT-4,llama3,llm_survey_zhao_23,guo2025deepseek_r1}. 
Despite their strengths, LLMs are often limited by the information contained in their training data and may struggle with answering time-sensitive or domain-specific queries accurately \cite{rishi2022survey_foundation,zhao2023survey_llm}.
In addition, they are prone to producing factually incorrect or ungrounded outputs, commonly referred to as hallucinations \cite{hallucination_survey23_zhang,huang2025survey_hallucination}. 
To alleviate this limitation, retrieval-augmented generation (RAG) \cite{neurips20_lewis_rag} has emerged as one of the most representative advanced techniques, aiming to enhance LLMs-based generation by integrating information retrieved from external data sources \cite{neurips20_lewis_rag,RAG_survey_Fan_KDD24,rag_survey24_huang,zeng_NAACL24_rag}. 
RAG has been applied to various fields, such as question answering \cite{rag_QA_Guu_ICML20,RAG_QA_tois}, recommendation systems \cite{RAG_recom_wu_kdd24,RAG_recom_Lu_ACL21}, and fact checking \cite{rag_fact_checking_Izacard_23,subquestion_emnlp22_chen}. 
For fake news detection, some studies \cite{folk_emnlp23_wang_and_shu,aacl23_zhang_gao_hiss,naacl24_chen} typically retrieve relevant information about a claim from external sources and then use LLMs for reasoning to determine the veracity of the claim. The reasoning process of LLMs can naturally be regarded as a form of explanation. 
However, the correctness of such reasoning strongly depends on the quality and credibility of the retrieved evidence, but the time-sensitive nature of breaking news makes it highly challenging to obtain reliable external knowledge. 
Existing approaches differ in their choice of retrieval sources. 
Some retrieve from verified knowledge bases like Wikipedia \cite{naacl24_yue_rag,acl2024_yue,folk_emnlp23_wang_and_shu,aacl23_zhang_gao_hiss}, which ensures high credibility but is limited in its coverage of up-to-date information for verifying breaking news. 
Others retrieve from unverified sources such as online comments or user-generated content \cite{naacl24_chen,factllama,zhang_coling24,EFND_TOIS}, which are more up-to-date but may contain false or biased information. Directly regarding them as gold evidence may lead to incorrect predictions. 
Thus, it remains a key challenge to accurately leverage the rich yet unverified information expressed in raw reports to support fake news detection.

\begin{figure}
    \centering
    \includegraphics[width=1\linewidth]{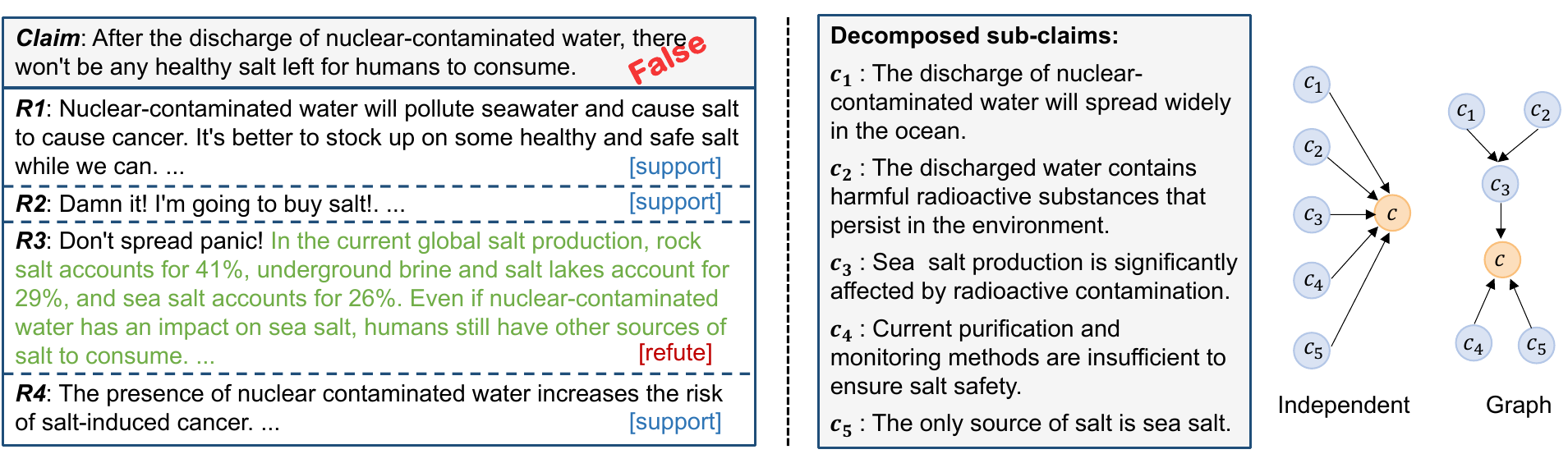}
    \caption{A false claim from Sina Weibo. Left: The comparison of informativeness and soundness between two competing parties serves as an indicator of veracity. Right: The different modeling of decomposed sub-claims. }
    \label{fig:intro}
\end{figure}

To address the challenge, we draw inspiration from recent advancements in stance detection \cite{jin_AAAI16, ma_WWW18,ruichao_sigir22,stance_rumor_tois24}, which suggest that the diverse and sometimes conflicting insights presented in raw reports are vital signals in the pursuit of truth.
Based on this insight, we propose to split the wisdom extracted from the retrieved raw reports into two distinct parties, which allows the detection to rely on the quality of wisdom rather than its quantity. 
For instance, as shown in Figure \ref{fig:intro} (left), a representative example illustrates how competing parties emerge around a false claim. 
For the supporting party, R1 and R4 briefly discuss the potential link between salt consumption and cancer, and R2 restates the claim without introducing new information. In contrast, R3 provides the refuting party with detailed evidence that illustrates its unique viewpoint, which is informative, sound, and persuasive. 
This observation motivates an assumption: in a pair of competing parties, the side that is more consistent with the actual veracity of the claim tends to be supported by evidence with greater informativeness and soundness\footnote{Soundness reflects the factual accuracy and logical coherence of the explanations \cite{exp-eval-metric_IJCAI23_Wang}.} than the opposing side. 
Accordingly, when an LLM is prompted to generate explanations based on the evidence from each side, the resulting explanations tend to reflect the difference in evidence quality. The explanation grounded in the side consistent with the true veracity label is more likely to exhibit higher quality.
We empirically examine this tendency in Appendix \ref{app:validation}. 
Consequently, the veracity of news can be determined through a comparative analysis between two competing parties. 
Therefore, our previous work \cite{wang_www24} introduced an LLM-equipped defense-based\footnote{The term ``defense'' in this paper denotes a \textit{debate-style reasoning process} between competing viewpoints to assess claim veracity. } explainable fake news detection (L-Defense) framework that models this comparative reasoning process to assess veracity. 
Specifically, L-Defense extracts two competing evidence sets, prompts an LLM to generate justifications by inferring reasons for both possible veracity labels, and compares their quality to determine the final verdict. The final explanation is then adaptively selected based on the prediction.

Despite the effectiveness of L-Defense in detecting fake news and generating human-friendly explanations, its outputs remain coarse-grained, especially for complex claims. 
For example, given the claim shown in Figure \ref{fig:intro} (left), L-Defense generates the explanation based on one retrieved report R3, which focuses solely on one aspect of the claim, i.e., ``\textit{sea salt is not the only available salt source}''. However, it fails to address other crucial aspects, such as ``\textit{whether the nuclear-contaminated water is actually being discharged?}'' and ``\textit{whether it affects the production of sea salt?}''. 
This limitation arises because retrieval and reasoning are performed at the full-claim level, which may overlook important sub-aspects and result in incomplete explanations that hinder user understanding.
To address this issue, several prior works propose to decompose a claim into multiple sub-claims to facilitate multi-step reasoning \cite{folk_emnlp23_wang_and_shu,subquestion_emnlp22_chen,aacl23_zhang_gao_hiss,naacl24_chen}. 
For each sub-claim, they retrieve relevant evidence and verify its veracity independently. As illustrated in Figure~\ref{fig:intro} (right, Independent), they aggregate the results to assess the overall claim. While this improves the granularity of reasoning and explanation, these methods do not consider the relationships between sub-claims. 
In many cases, inferring the veracity of one sub-claim may require others as contexts. 
Modeling the claim and sub-claims in a graph structure will further improve reasoning. For example, as shown in Figure~\ref{fig:intro} (right, Graph), the sub-claim $c_3$ (``\textit{sea salt production is affected}'') depends on $c_1$ (``\textit{nuclear-water spreads widely}'') and $c_2$ (``\textit{the discharged substances persist in the environment}''). Such dependencies are essential for accurate reasoning, and overlooking them can lead to fragmented explanations. Moreover, unlike the approaches that treat sub-claims independently, the graph clearly indicates that the claim $c$ relies on $c_3$, $c_4$, and $c_5$ simultaneously, providing a more intuitive explanation. 
Based on this insight, we enhance L-Defense with claim decomposition and propose modeling the sub-claims as a graph structure, which enables the representation of rich dependency relationships among sub-claims.

To this end, in this study, we propose a novel graph-enhanced defense framework for explainable fake news detection, which strives to perform fine-grained reasoning based on claim-centered graphs and pursue the veracity of claims in a defense-like way. 
Specifically, we first introduce a graph construction module that decomposes a news claim into several sub-claims and derives the dependency edges between them using an LLM, resulting in a claim-centered graph. 
Then, for each sub-claim, we use RAG to retrieve its relevant evidence and prompt the LLM to generate competing explanations by inferring reasons towards two possible veracity labels. 
This step ensures that diverse perspectives from raw reports are fully considered. 
To make the final prediction, we integrate the generated explanations into the claim-centered graph and employ an LLM to conduct reasoning over the graph, naturally combining graph-based reasoning with defense-like inference. In this way, comparing the informativeness and soundness of the two explanations helps the model determine the most plausible veracity label. 
Finally, based on the predicted label and the constructed claim-centered graph, we prompt an LLM to infer the veracity of each sub-claim and generate a concise textual explanation summarizing the reasoning. Meanwhile, we can obtain a subgraph as the final explanation graph by removing the explanations that are not consistent with the sub-claims' veracities. This explanation graph highlights the dependencies among all relevant sub-claims, their veracities, and corresponding explanations, enabling users to intuitively understand the truthfulness of each aspect of the claim. 
Our main contributions are summarized as:
\begin{itemize}
    \item[---] We substantially extend the LLM-equipped defense-based explainable fake news detection (L-Defense) method published in our recent work \cite{wang_www24} by mimicking the human thinking process. Instead of assessing a claim as a whole, we decompose it into sub-claims, enabling the model to handle more complex claims and support more fine-grained reasoning.
    \item[---] We develop a novel graph-enhanced defense framework that not only uses competing wisdom extracted from retrieved reports, but also leverages the internal dependencies among sub-claims, enabling more structured and reliable reasoning for fake news detection.
    \item[---] The proposed framework generates an intuitive explanation graph accompanied by fine-grained textual explanations, which are comparable to those of human experts and do not require supervision from debunked reports.
    \item[---] We conduct extended experiments by exploring multiple variants of G-Defense, which validate both the effectiveness and the extensibility of the proposed framework.
    \item[---] The proposed framework achieves state-of-the-art performance on two benchmarks while providing more comprehensive and reliable veracity explanations.
\end{itemize}

\section{Related Work} \label{sec:related_work}
\subsection{Explainable Fake News Detection Before the LLM Era}
Explainability is a critical challenge in fake news detection, and many studies have been devoted to addressing it. 
Early works typically use attention mechanisms to highlight salient phrases \cite{DeClarE_EMNLP18,conflict_wu_aaai21}, news attributes \cite{XFake_yang_www19}, and suspicious users \cite{GCAN_Lu_and_Li_ACL20,FND_tois} to provide explainability. In order to gain more human-readable explanations, later works further employ sentence-level attention mechanisms to identify important sentences as explanations \cite{Nie_AAAI19,HAN_ma_acl19,dEFEND_KDD19}. However, these methods merely uncover regions with high contributions to the veracity prediction rather than regarding the explanation generation as a dependent task.
To address it, \citet{EXTABS_EMNLP20_Kotonya} views the explanation generation task as a pre-trained extractive-abstractive summarization task. \citet{GenFE_ACL20} formats explainable fake news detection as a multi-task learning problem and develops a joint model trained using external debunked reports gathered from fact-checking websites. 
However, debunking claims is a labor-intensive and time-consuming process, leading to significant limitations in coverage and delays in the availability of debunked journalism. 
To alleviate this issue, \citet{zhiwei22coling} employs debunked reports as supervision during training, and focuses on the majority opinions expressed in relevant crowdsourced raw reports to enhance prediction and evidence extraction. However, it overlooks the inaccurate and biased information in unverified reports, which may lead to misleading results. Moreover, the extracted evidence, being discrete and fragmented, still struggled to meet users' needs for comprehensive and intuitive explanations.
In addition, all of these methods struggle with complex news claims, as they directly analyze the claim without digging out fine-grained relevant information. 
Fortunately, the emergence of LLMs with powerful reasoning and generating abilities presents promising solutions.

\subsection{Large Language Models in Fake News Detection} \label{sec:related_LLM}

Recent advances in LLMs have shown remarkable performance in various tasks. Researchers have attempted to employ LLMs for fake news detection. With pre-training on massive corpora, LLMs inherently contain lots of commonsense and factual knowledge in their parameters, which can be beneficial for identifying false information. 
\citet{aaai24_hu} explores the potential of LLMs in fake news detection, and points out that they are appropriate to provide multi-perspective rationales for detection, but underperform the fine-tuned language models on detection performance. Then they propose an adaptive rationale guidance network in which fine-tuned language models selectively acquire insights on news analysis from the LLMs’ rationales. 
Similarly, some works prompt LLMs to generate more viewpoints or human-like reviews to provide more information. 
DELL \cite{acl24_findings_wan} engages LLMs to generate news reactions to represent diverse perspectives and explanations for proxy tasks to enrich the contexts of news articles. With an LLM-based expert ensemble module, it selectively integrates LLM-generated contents to obtain the final prediction. 
GenFEND \cite{cikm24_nan} generates comments by prompting LLMs with diverse user profiles and analyzes them from multiple perspectives. 
Although their success in providing multiple viewpoints, LLMs suffer from severe hallucinations \cite{hallucination_survey23_ji,hallucination_survey23_zhang,hallucination_bang_arxiv23}, such contents generated through a role-play strategy inevitably introduce nonfactual or biased information. Moreover, LLMs rely on static knowledge encoded during pretraining, which limits their ability to handle breaking news. 

To address these limitations, RAG \cite{neurips20_lewis_rag,tacl23_ram_rag,langchain} has been proposed to enhance LLMs by incorporating externally retrieved relevant information during generation. 
It enables LLMs to produce accurate and contextually relevant outputs with augmented external knowledge, mitigating the hallucination issue, achieving great results on many knowledge-intensive tasks \cite{rag_survey23_gao,rag_survey24_huang}. 
In fake news detection, some works introduce RAG to improve the performance of detection. 
To counter misinformation, RARG \cite{naacl24_yue_rag} collects supporting evidence from scientific sources and generates responses based on the evidences via reinforcement learning from human feedback (RLHF). 
RAFTS \cite{acl2024_yue} retrieves relevant documents from verified sources like Wikipedia and conducts fact verification by synthesizing contrastive arguments.
Furthermore, to detect complex news claims, many works propose to first decompose the claim into several sub-claims or questions, and then verify these sub-claims step-by-step through retrieval to derive the final prediction. 
For example, FOLK \cite{folk_emnlp23_wang_and_shu} leverages LLMs to translate the claim into a First-Order-Logic (FOL) clause consisting of several sub-claims, and then retrieves knowledge-grounded answers from external knowledge sources for each sub-claim. Based on FOL-guided reasoning over a set of knowledge-grounded question-and-answer pairs, it makes veracity predictions and generates explanations. 
\citet{aacl23_zhang_gao_hiss} introduces a hierarchical step-by-step prompting method that directs LLMs to separate a claim into several sub-claims and then verify each of them via multiple question-answering steps progressively.

Despite their success in detection and explanation generation, they are developed for an unrealistic scenario. They rely on knowledge from verified sources, which is often limited for emerging news claims, or access to evidences published after a claim was verified, suffering from ground-truth leakage \cite{naacl24_chen}. 
To address this issue, \citet{naacl24_chen} introduces a pipeline for realistic fact-checking of complex claims. It decomposes claims into sub-questions and answers them by retrieving raw evidence from web documents. 
\citet{factllama} uses LoRA tuning \cite{lora_ICLR22} to train a LLaMA-based \cite{llama} detector with filtered retrieved knowledge.
\citet{zhang_coling24} propose a two-level strategy to gather fine-grained feedback from the LLM, which serves as a reward for optimizing the retrieval policy, by rating the retrieved raw documents based on the non-retrieval ground truth. 
Although these works aim to operate in more realistic settings, raw documents often contain inaccurate or biased information, which may lead to incorrect predictions. 
To address the challenges of incorporating such unverified and potentially misleading evidence in realistic settings, we propose a defense-like strategy that helps the model focus on more reliable information. 
In addition, these methods demonstrate that claim decomposition could bring great performance improvement, as also supported by the detailed analysis in \citet{hu_naacl25_analysis_claimdecomposition}. 
However, they typically verify sub-claims independently and aggregate the results to assess the overall claim, overlooking the underlying dependencies and interactions among them. To address this limitation, we propose constructing a claim-centered graph that explicitly captures the relationships between sub-claims and facilitates more comprehensive reasoning.

\subsection{Large Language Models on Graphs} \label{sec:related_GLM}
While LLMs are widely applied to textual data, an increasing number of studies explore their use in scenarios where textual information is accompanied by structural data in the form of graphs. Specifically, many works have extended LLMs to handle text-rich graphs \cite{kdd23_chen_llm_graph,eacl24_findings_ye_instructGLM,iclr24_fatemi_graph-encoding-for-llm,tkde24_jin_survey_LLM-on-Graph}. A straightforward method to adapt LLMs to graphs is to translate the graph structure into a text sequence with specially designed rules and then regard the LLM as a predictor to finish the graph-related task. For example, InstructGLM \cite{eacl24_findings_ye_instructGLM} uses natural language to describe graph structure for each node and then instructs an LLM for node classification and link prediction. \citet{arxiv23_huang_glm} investigates why the structure information in text format can improve the performance of LLM on node classification and finds that the structure information is beneficial when the textual information associated with the node is inadequate. \citet{iclr24_fatemi_graph-encoding-for-llm} explores various graph-to-text encoding strategies and observes that the simple description of a node's one-hop index information yields strong performance across multiple tasks. Inspired by these works, we propose to engage an LLM to conduct veracity prediction on our generated claim-centered graph to effectively utilize both textual and structural information. 

\section{Proposed Approach} \label{sec:approach}

\begin{figure*}[t]
    \centering
    \includegraphics[width=1\textwidth]{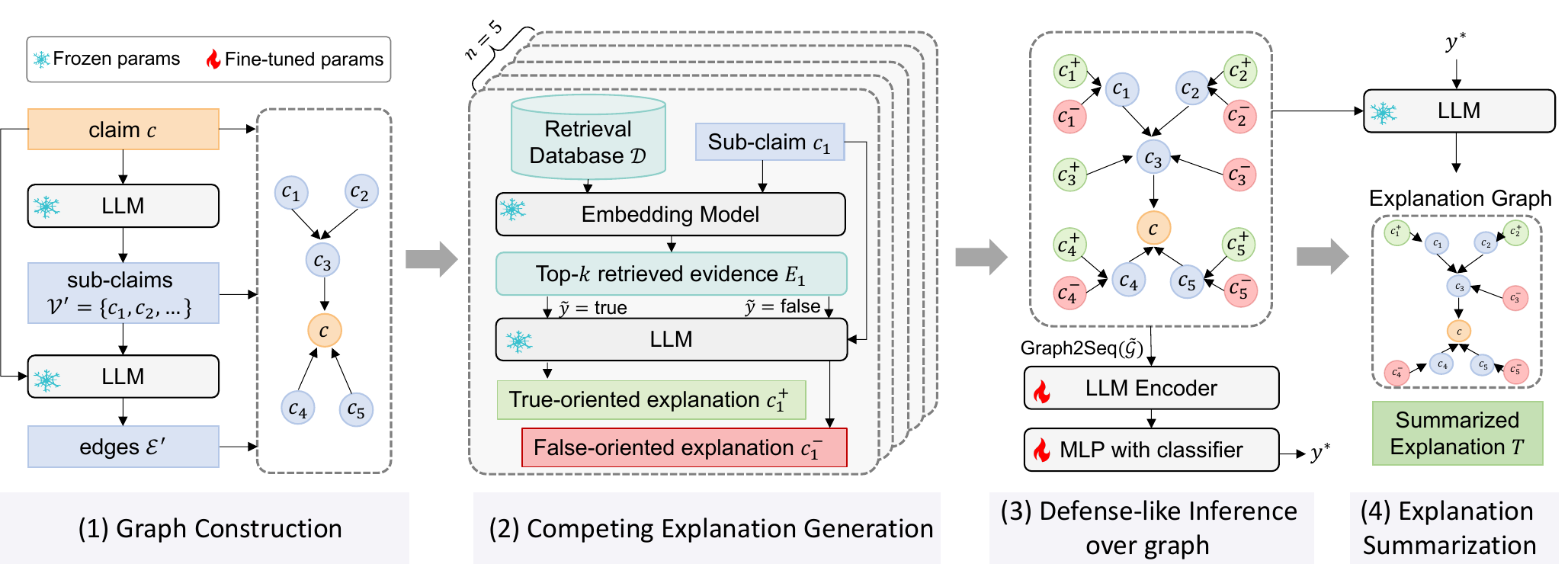}
	\caption{An architecture overview of the proposed \textbf{G}raph-enhanced \textbf{Defense} framework for explainable fake news detection (\textbf{G-Defense}). Given a claim $c$, the graph construction module first constructs a claim-centered graph through claims decomposition and edge generation, where the claim is decomposed into $n$ sub-claims. Each sub-claim $c_i$ is then individually processed to retrieve evidence and generate competing explanations. Next, the claim-centered graph enriched with explanations is used for defense-like inference to derive the veracity prediction of the claim. Finally, an explanation graph and a summarized explanation $T$ can be derived from the last module. 
	}
	\label{fig:model_architecture} 
\end{figure*}

This section begins with the task definition for explainable fake news detection and an overview of the proposed \textbf{G}raph-enhanced \textbf{Defense} framework (\textbf{G-Defense}). 
It subsequently presents its four modules: the graph construction module (\S \ref{subsec:graph_construction}), the competing explanation generation module (\S\ref{subsec:explanation_gen}), the defense-like inference module (\S\ref{subsec:glm_inference}), and the explanation summarization module (\S\ref{subsec:explanation_sum}).

\paragraph{Task Definition. }
Given a news claim $c$ associated with a gold veracity label $y$, explainable fake news detection aims to predict a veracity label $y^*$ of claim $c$ and provide an explanation supporting the prediction. 
To assist this process, a set of relevant raw reports $\gD = \{d_1, d_2, ..., d_{|\gD|}\}$ are pre-retrieved as external database to support reasoning and explanation generation, where each $d_j = \{x_1, x_2, ..., x_{|d_j|}\}$ denotes a relevant report consisting of a sequence of sentences. 
Notably, the reports from fact-checking sources and any content published after the release of corresponding fact-check conclusions are strictly excluded from $\gD$ to avoid potential leakage of ground truth. 
In this work, we evaluate our framework on two benchmark datasets, \textsc{RAWFC} and \textsc{LIAR-RAW}~\cite{zhiwei22coling}, each pairing news claims with raw reports retrieved before the release of official fact-checking articles. 
These datasets provide a realistic setting for early fake news detection, where only unverified reports from the open web are available. 
Depending on the dataset, the predicted label $y^*$ can be either in the set $\{\textit{false}, \textit{half}, \textit{true}\}$ for \textsc{RAWFC}, or in the finer-grained set $\{\textit{pants-fire}, \textit{false}, \textit{barely-true}, \textit{half-true}, \textit{mostly-true}, \textit{true}\}$ for \textsc{LIAR-RAW}. 
Detailed descriptions and statistics for both datasets are provided in \S\ref{subsec:exp_setup}.

\paragraph{Overview of G-Defense. }
G-Defense decomposes a claim into several sub-claims and organizes them into a dependency graph, enabling fine-grained reasoning. Moreover, for each sub-claim, G-Defense generates supporting and refuting explanations from retrieved raw reports, allowing the model to compare competing viewpoints before making a final decision. 
The framework shown in Figure \ref{fig:model_architecture} consists of four modules: 
(1) graph construction, the claim is decomposed into sub-claims and organized into a dependency graph;
(2) competing explanation generation, supporting and refuting explanations are produced for each sub-claim using retrieved raw reports;
(3) defense-like inference, the explanation-enriched graph is encoded and used to predict the veracity;
(4) explanation summarization, a final explanation graph and concise textual justification are generated to provide intuitive explanations.

\subsection{Graph Construction} \label{subsec:graph_construction}

In real-world scenarios, news claims are often complex and multifaceted, containing multiple implicit assumptions and logical links. Human experts typically do not evaluate such claims as a whole. Instead, they tend to break them down into smaller pieces, evaluate each sub-claim using background knowledge and evidence, and then form a global decision based on how these pieces interact. 
Inspired by this human-like thinking process, prior works have proposed to decompose news claims into multiple sub-claims \cite{folk_emnlp23_wang_and_shu,subquestion_emnlp22_chen,aacl23_zhang_gao_hiss,naacl24_chen}. However, these methods treat sub-claims as independent units, overlooking the interactions between them. 
Inferring the veracity of one sub-claim sometimes requires some additional context, such as the interpretation of other sub-claim(s). Some sub-claims may only be meaningful when referring to others, so treating them as isolated units may lead to fragmented or oversimplified reasoning.  

\begin{figure*}[t]
    \centering
    \includegraphics[width=1\textwidth]{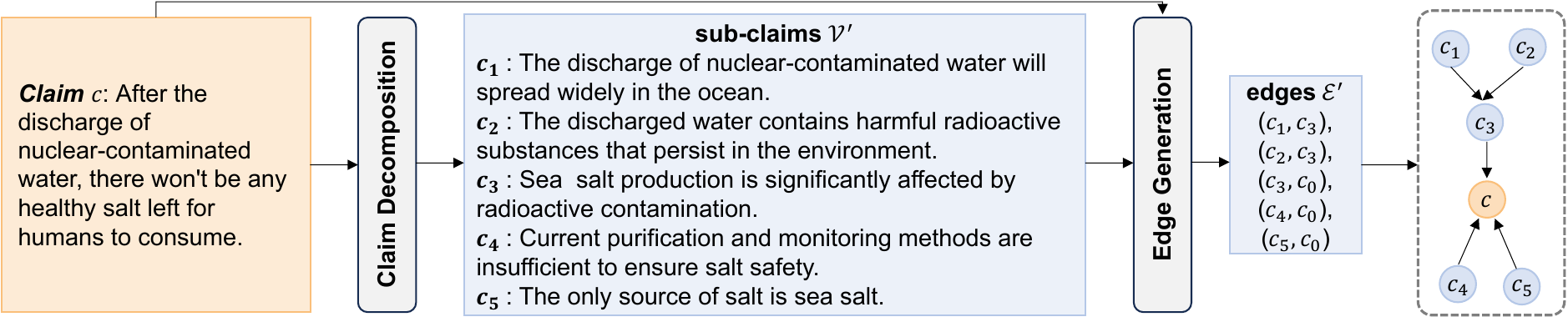}
	\caption{An illustrative example of the graph construction process.}
    \label{fig:graph_construction}
\end{figure*}

To address the above issue, we propose to model the dependencies among sub-claims by constructing a claim-centered graph over the decomposed sub-claims. In this graph, nodes represent the original claim and its sub-claims, while edges capture their dependencies. This structured representation provides a solid foundation for downstream reasoning and explanation generation. 
To construct the graph, we design two steps: claim decomposition and edge generation. 
An example of the graph construction process is shown in Figure \ref{fig:graph_construction}.
 
\paragraph{Claim Decomposition}
We leverage an LLM to capture all explicit aspects in the given claim $c$ that will affect its veracity and output the decomposed sub-claims. The process can be formulated as:
\begin{align}
    \text{LLM}(I^{v}(c))=\gV'=\{c_1,c_2,...,c_n\}, \label{eq:claim_decomposition}
\end{align}
where $I^{v}$ denotes the prompt used to guide the decomposition and the detail of $I^{v}$ is listed in Appendix \ref{app:prompt}, $n$ is the number of decomposed sub-claims, and $\gV'$ denotes the set of sub-claims. We then define the node set of the claim-centered graph $\gG$ as:
\begin{align}
    \gV=\gV' \cup \{c\}. \label{eq:claims_node_set}
\end{align}

\paragraph{Edge Generation} After the claim decomposition, to capture the potential dependency entailed in these sub-claims and claim, we further instruct an LLM to generate dependency edges between nodes:
\begin{align}
    \text{LLM}(I^{e}(\gV))=\gE'=\{e_1,e_2,...,e_m\}, \label{eq:edge_generation}
\end{align}
where $I^{e}$ denotes the instruction prompt used for edge generation and the detail of $I^{e}$ is in Appendix \ref{app:prompt}, and each edge $e=(c_i, c_j)$ is a directed edge from source node $c_i$ to target node $c_j$. 
Note that the claim $c$ is also included in the input and indexed as $c_0$ so that the LLM has access to the complete set of nodes when inferring potential dependencies.
Since each sub-claim describes an aspect of the claim, we additionally include an edge from every sub-claim to $c$ to reflect this inherent relation in the final graph.
The final edge set is defined as:
\begin{align}
    \gE=\gE' \cup \{(c_i, c) \mid i = 1, 2, \dots, n\}. \label{eq:edge_set}
\end{align}

After claim decomposition and edge generation, we obtain the claim-centered graph $\gG=(\gV, \gE)$, which can be used for the sequential reasoning and explanation generation.

\subsection{Competing Explanation Generation with RAG} \label{subsec:explanation_gen}

Relying on evidence from a single perspective may lead to inaccurate or biased reasoning, especially when the available information is noisy or partial. 
Considering multiple viewpoints provides a more comprehensive analysis of a claim's veracity. 
We assume that the difference between supporting and refuting explanations, in terms of informativeness and soundness, offers important signals for downstream inference. 
For example, if one explanation is significantly more detailed, logically sound, and better supported by evidence than its opposing party, it can steer the model to make a more accurate judgment of the claim's veracity.

\begin{figure*}[t]
    \centering
    \includegraphics[width=1\textwidth]{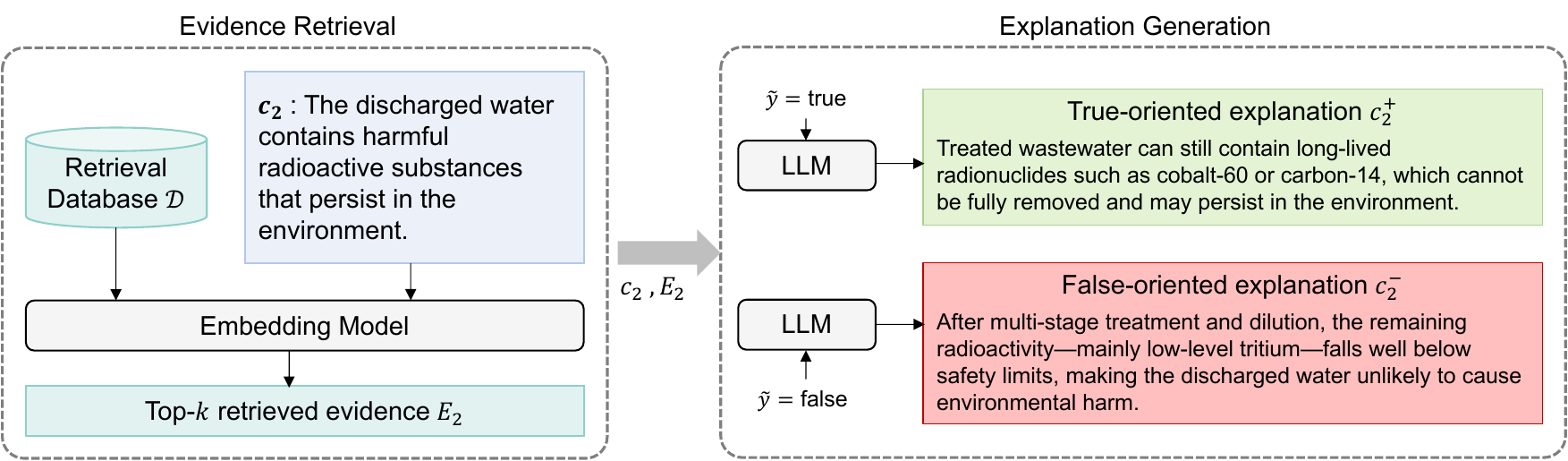}
	\caption{An illustrative example of the competing explanation generation process for a sub-claim.}
	\label{fig:explanation_generation}
\end{figure*}

To support the generation of such contrasting explanations, we adopt the retrieval-augmented generation technique. Specifically, for each sub-claim, we retrieve a set of relevant evidences from externally collected raw reports based on cosine similarity between their embedding representations.  
These retrieved evidences are used as the input context to the LLM, prompting it to generate two competing explanations: one supporting the sub-claim and the other refuting it.  
This retrieval process ensures that the LLM can ground its reasoning in diverse real-world perspectives, which is particularly important when no verified knowledge base is available. 
An example of the process of graph construction is shown in Figure \ref{fig:explanation_generation}.

\paragraph{Evidence Retrieval}
Since a report may contain evidences related to different sub-claims, we first split each report into individual sentences, resulting in a corpus of candidate evidential sentences $\gX = \{x_j\}_{j=1}^o$, where $o = \sum_{d \in \gD} |d|$. 
We thereby adopt a vanilla pre-trained transformer encoder \cite{transformer} as the embedding model to generate representations for each sub-claim and candidate evidence. 
Specifically, given a sub-claim $c_i$ or a candidate evidence $x_j$, their representations are derived from the Transformer-Encoder as:
\begin{align}
    \vh_{c_i} &= \pool(\transformerenc(c_i)),\\
    \vh_{x_j} &= \pool(\transformerenc(x_j)).
\end{align}
We then use cosine similarity to retrieve relevant evidence sentences for each sub-claim, as it provides a simple yet effective way to measure the semantic relevance between embeddings. For each sub-claim, the top-$k$ most relevant candidates are selected as the final evidence set. Formally, the final evidence set for each sub-claim is:
\begin{align}
    E_i = \{x_j|\text{cos}(\vh_{c_i}, \vh_{x_j})~\text{is in the top-}k \}. \label{eq:evidences_set}
\end{align}

Compared with retrieving evidence for the entire claim, retrieving evidence separately for each sub-claim ensures that all aspects of the claim are supported by the most relevant evidence, rather than being dominated by generic parts. Therefore, more diverse evidence can be covered, further enhancing explanation generation and downstream inference.

\paragraph{Explanation Generation}
To effectively leverage the rich information contained in retrieved evidences and obtain coherent explanations, we leverage an LLM to conduct abductive reasoning to explain why a sub-claim $c_i$ should be assigned the veracity label $\tilde y$ based on the retrieved evidence sets $E_i$ and a given prior label $\tilde y$, which can be formulated as:
\begin{align}
    c_i^v = \text{LLM}(I^r(c_i, E_i, \tilde y)), \label{eq:competing_explanation}
\end{align}
where the prior label $\tilde y$ can be false or true, determines whether the veracity-oriented explanation $c_i^v$ corresponds to a false-oriented explanation $c_i^-$ or a true-oriented explanation $c_i^+$. $I^r$ is the curated template to integrate these information and its details are as follows. We prompt the LLM with $I^r$ to generate an explanation $c_i^{-/+}$ that reasons about how to infer the veracity label $\tilde y$ based on the claim $c_i$ and the evidence set $E_i$. 

\begin{tcolorbox}[title={Prompt used for veracity-oriented explanation generation, $I^{r}$},
coltitle=black, colback=gray!10, colframe=gray!30, boxrule=0.5mm, arc=2mm, left=2mm, right=2mm]
Given a claim: $c_i$, a veracity label $\tilde y$, please give me a streamlined rationale associated with the claim, without explicitly indicating the label, for how it is reasoned as $\tilde y$. Below are some sentences that may be helpful for the rationale, but they are mixed with noise: $E_i$. \\
Note, please do not repeat the claim and the label in your explanation, just directly output your streamlined rationale in a short and clear manner.
\end{tcolorbox}

The reasoning is performed for both \textit{false} and \textit{true} and thus two reasoning texts are obtained. 
As detailed previously, the evidence that is consistent with the actual veracity of the claim brings more information and is more reasonable than the competing one. Thus, the LLM prefers to generate solid reasoning in favor of it, while providing weak reasoning with inaccurate information for its competitor, as demonstrated in Appendix \ref{app:validation}. 
In this way, the two LLM-generated veracity-oriented reasoning $c_i^-$ and $c_i^+$, which can be viewed as two explanations to clarify their corresponding veracity labels, will always possess a relative strength in confidence, greatly facilitating the detection of fake news.

\subsection{Defense-like Inference over Graph}   \label{subsec:glm_inference}

With the claim-centered graph and two veracity-oriented explanations for each sub-claim derived from LLM, we develop a defense-based fake news detector. This detector aims to identify the relative strength of the two explanations from their defense and analyze the dependency relationships between sub-claims and claims ultimately providing the veracity. 
To effectively make use of the competing explanations, we view a pair of explanations as two property nodes of their corresponding sub-claim node and add them into the graph. Formally, the final defense-augmented claim-centered graph is like:
\begin{align}
     \tilde \gG &= (\tilde \gV, ~\tilde \gE),\label{eq:full_graph}\\ 
     \tilde \gV &= \gV \cup \{c_i^-, c_i^+\}_{i=1}^n, \label{eq:full_node} \\ 
     \tilde \gE &= \gE \cup \{( c_i^-, c_i), (c_i^+, c_i)\}_{i=1}^n.
\end{align}
Then, we serialize the claim-centered graph $\tilde \gG$ into a sequence $\rmH_{\tilde \gG}$ as below:
\begin{align}
    \rmH_{\tilde \gG} &= \text{Graph2Seq}(\tilde \gG), \label{eq:graph2seq}
\end{align}
where $\text{Graph2Seq}(\cdot)$ denotes a template to describe the graph structure with natural language. Here, we use the most straightforward way to translate, whose effectiveness has been demonstrated by many previous works \cite{iclr24_fatemi_graph-encoding-for-llm}. 
We first point out all the nodes in the graph, and then describe the directed edges connected to each node in sequence. The claim is indexed as node 0, and the sub-claims are indexed starting from node 1. 
For example, given a graph $\tilde \gG_t$ of a claim $c$ and its five decomposed sub-claims as shown in Figure \ref{fig:model_architecture}, $\rmH_{\tilde \gG_t}=\text{Graph2Seq}(\tilde \gG_t)=$ \textit{``Directed Graph describes a graph among 0, 1, 2, 3, 4, 5. Node 0 is connected to nodes 3, 4, and 5 by incoming edges. Node 3 is connected to nodes 1 and 2 by incoming edges.'' } 
Notably, the two veracity-oriented explanations $c_i^{-/+}$ of each sub-claim are not explicitly included as separate nodes in the $\text{Graph2Seq}(\tilde \gG_t)$. This is because the explanations are treated as attributes associated with each sub-claim node, providing supplementary reasoning information rather than altering the structural relationship within the graph. Consequently, in the serialization process, we mainly focus on describing the claim, the sub-claims, and their dependency edges. 
This sequence $\rmH_{\tilde \gG}$ is fed into the LLM-Encoder, along with the contents of the original claim, its sub-claims, and a classification query $Q$, for final veracity prediction, which can be formulated as: 
\begin{align}
    \vh_{cls} &= \text{LLM-Enc}(I^a(\rmH_{\tilde \gG}, \tilde \gV, Q)), \label{eq:graph_encoding}
\end{align}
where $\vh_{cls}$ is the contextual representation of the query and the claim-centered graph, which jointly encode both textual and structural information. $I^a$ denotes the prompt template used to integrate all input components and its content is as below. The content of the classification query $Q$ can be found in $I^a$. 
\begin{tcolorbox}[title={Prompt used for aggregating all information into the LLM-Encoder, $I^{a}$},
coltitle=black, colback=gray!10, colframe=gray!30, boxrule=0.5mm, arc=2mm, left=2mm, right=2mm]
Graph used for fake news detection: \\
\# \textbf{Node Content}: \\
Node 0 (claim): $c$\\
Node 1 (Sub-claim n): $c_1$; \\
Competing explanations: True-oriented explanation: $c_1^+$; False-oriented explanation: $c_1^-$.\\
...\\
Node $n$ (Sub-claim $n$): $c_n$;\\
Competing explanations: True-oriented explanation: $c_n^+$; False-oriented explanation: $c_n^-$.\\
\# \textbf{Graph Structure}: $\rmH_{\tilde \gG}$\\
\# \textbf{Query ($Q$)}: What is the label of Node 0 (claim)? Please directly output your predicted label from \{false, half, true\}/\{pants-fire, false, barely-true, half-true, mostly-true, true\}.

\end{tcolorbox}

Then, we define a classifier on top of the rich representation for veracity prediction,
\begin{align}
    \vp = P(z|\vh_{cls}) \triangleq \softmax (\mlp (\vh_{cls})).
\end{align}

The training objective of the detection task is written as:
\begin{align}
    \gL = - \sum\nolimits_\gD \log \vp_{[\hat y = y]}, \label{eq:veracity_loss}
\end{align}
where $\vp_{[\hat y = y]}$ denotes fetching the probability value corresponding to the veracity label $y$. 

The inference procedure can be simply written as:
\begin{align}
    y^* = \arg \max \vp. \label{eq:label_prediction}
\end{align}

\begin{algorithm}[t]
\caption{G-Defense Inference Algorithm}
\label{alg:g-defense}
\begin{flushleft}
\textbf{Input:}  News claim $c$, External Database $\mathcal{D}$, Fine-tuned LLM-Encoder \\
\textbf{Output:} Predicted label $y^*$, Explanation graph $G^*$, Summarized explanation $T$
\end{flushleft}
\begin{algorithmic}[1]
\Statex\textcolor{gray}{// Stage 1: Graph Construction}
\State Claim decomposition: $V' \gets $ Equation (\ref{eq:claim_decomposition})
\State Edge generation: $\gE' \gets$ Equation (\ref{eq:edge_generation})
\Statex\textcolor{gray}{// Stage 2: Competing Explanation Generation with RAG}
\For{each sub-claim $c_i \in V'$}
    \State Evidence retrieval: $E_i \gets $ Equation (\ref{eq:evidences_set})
    \State Explanation generation: $c_i^{-/+} \gets $ Equation (\ref{eq:competing_explanation})
\EndFor
\Statex\textcolor{gray}{// Stage 3: Defense-like Inference over Graph}
\State Obtain claim-centered graph: $\tilde \gG \gets$ Equation (\ref{eq:full_graph})
\State Graph2Seq: $\rmH_{\tilde \gG} \gets$ Equation (\ref{eq:graph2seq})
\State Obtain the embedding for classification: $h_{cls} \gets $ Equation (\ref{eq:graph_encoding}) 
\State Get prediction result: $y^* \gets$ Equation (\ref{eq:label_prediction})
\Statex\textcolor{gray}{// Stage 4: Explanation Summarization}
\State Obtain explanation graph and summarized explanation: $\gG^*, T \gets$ Equation (\ref{eq:final_explanation})
\State \Return $y^*, \gG^*, T$
\end{algorithmic}
\end{algorithm}

\subsection{Explanation Summarization} \label{subsec:explanation_sum}
Based on the derived claim-centered graph and the predicted veracity, we further ask the LLM to give a final summarized explanation. We prompt the LLM to judge the veracity of each sub-claim based on the predicted veracity of the claim and its competing explanations, and instruct the LLM to generate a concise explanation for the full claim. The used prompt $I^{s}$ to instruct the LLM is listed in Appendix \ref{app:prompt}. The process can be formulated as:
\begin{align}
    \gG^*, T = \text{LLM}(I^{s}(\rmH_{\tilde \gG}, \tilde \gV, y^*)). \label{eq:final_explanation}
\end{align}
In this way, the final explanation consists of two parts. The first part $\gG^*$ is provided in the form of a graph, where explanations inconsistent with the veracity of their associated sub-claims are removed from $\tilde{\gV}$ in Eq.(\ref{eq:full_node}). 
Note that this operation filters out only veracity-inconsistent explanations and does not affect the relevant contextual information about the sub-claim. Because both supporting and refuting explanations are generated from the same retrieved evidence, contextual sentences typically appear in both. These necessary background details therefore remain available regardless of which evidential explanation is removed. 
The second part $T$ is a textual explanation for the claim, as illustrated on the right side of Figure \ref{fig:model_architecture}. 
Compared with providing only a plain textual explanation, the explanation graph $\gG^*$ explicitly visualizes the reasoning process by showing how individual sub-claims contribute to the final veracity prediction, making the explanation more transparent and easier to understand. Building on this, by presenting the dependencies and veracity labels of sub-claims in a structured format, the graph enables users to quickly understand which aspects of the claim are true or false. In general, the graph-based explanation offers a fine-grained and more intuitive view of the claim's veracity reasoning, complementing the textual summary $T$ and enhancing the overall explainability of the proposed framework. 
During the automatic evaluation of the quality of explanation, we convert the explanation graph into text using Graph2Seq($\cdot$) and additionally provide the full textual content of all nodes. 
The overall inference process of our G-Defense is summarized in Algorithm \ref{alg:g-defense}.

\section{Experiment} \label{sec:exp}

In this section, we first introduce the experimental setup in \S \ref{subsec:exp_setup} and then evaluate G-Defense on the task of fake news detection in \S\ref{subsec:eval_veracity}. 
To demonstrate the significance of each proposed module, we conduct an extensive ablation study as detailed in \S \ref{subsec:ablation}. 
Moreover, we evaluate the quality and usefulness of fine-grained explanations generated by G-Defense in \S \ref{subsec:eval_explanation}. 
In \S \ref{subsec:case_study}, we present a case study to demonstrate how our proposed framework improves performance and provides intuitive graph-based explanations. 
Finally, in \S \ref{subsec:error_propa_discussion}, we discuss the inherent robustness of G-Defense to error propagation. 

\subsection{Experimental Setup} \label{subsec:exp_setup}

\begin{table}[t]
    \centering
    \caption{Summary statistics of datasets. \# half-true* is also denoted as \# half in RAWFC. } 
    \begin{tabular}{lcc}
        \hline
        \textbf{Dataset} & \textbf{RAWFC} & \textbf{LIAR-RAW} \\
        \hline
        \multicolumn{3}{l}{Claim} \\
        \ \ \ \# Claims & 2,012 & 12,590 \\
        \ \ \ \# pants-fire & - & 1,013 \\
        \ \ \ \# false & 646 & 2,466 \\
        \ \ \ \# barely-true & - & 2,057 \\
        \ \ \ \# half-true* & 671 & 2,594 \\
        \ \ \ \# mostly-true & - & 2,439 \\
        \ \ \ \# true & 695 & 2,021 \\
        Veracity Labels & 3 & 6 \\
        \hline
        \multicolumn{3}{l}{Report per claim} \\
        \ \ \ \# min & 1 & 1 \\
        \ \ \ \# max & 30 & 30 \\
        \ \ \ \# avg & 21.0 & 12.3 \\
        \hline
        \multicolumn{3}{l}{Sentence per report} \\
        \ \ \ \# min & 1 & 1 \\
        \ \ \ \# max & 155 & 59 \\
        \ \ \ \# avg & 7.4 & 5.5 \\
        \hline
    \end{tabular}
    \label{tb:datasets}
\end{table}

\paragraph{Datasets. }
Benchmarks for explainable fake news detection, such as \textsc{PubHealth} \cite{EXTABS_EMNLP20_Kotonya}, rely on official debunked reports written by journalists as evidence. However, such datasets are impractical for real-world, timely fake news detection, as official reports are typically published long after fake news has spread. Also, it does not align with our proposed model, which aims to detect fake news using externally retrieved raw reports. 
Therefore, to evaluate the effectiveness of detecting fake news before official reports are available, we select RAWFC and LIAR-RAW \cite{zhiwei22coling} to assess our proposed method. 
These datasets better reflect the general challenge of fake news detection with explainability, as they use raw reports retrieved from the Internet. And any reports published after the claims were verified are filtered out, ensuring a more realistic evaluation setting.

RAWFC contains the claims collected from Snopes\footnote{\url{https://www.snopes.com/}} and relevant raw reports by retrieving claim keywords. 
LIAR-RAW is extended from the public dataset LIAR-PLUS \cite{LIAR-PLUS_Alhindi_18} with retrieved relevant raw reports, containing claims from Politifact\footnote{\url{https://www.politifact.com/}}. 
Both datasets originate from fact-checking platforms, i.e., Snopes and PolitiFact, which collect and verify claims that initially circulated on social media and online news outlets. 
In RAWFC, claims are categorized into three labels: \{\textit{false}, \textit{half}, \textit{true}\}, while those in LIAR-RAW are annotated with six veracity labels: \{\textit{pants-fire}, \textit{false}, \textit{barely-true}, \textit{half-true}, \textit{mostly-true}, \textit{true}\}. 
Both datasets follow an 8:1:1 split for train/valid/test sets. 
The detailed statistics for these datasets are listed in Table \ref{tb:datasets}.

RAWFC has a simpler veracity classification scheme with only three labels, whereas LIAR-RAW adopts a finer-grained six-label classification. This difference makes LIAR-RAW more nuanced in distinguishing varying degrees of truthfulness, while RAWFC provides a more generalized and easily interpretable classification. 
Moreover, RAWFC provides significantly more supporting raw reports per claim than LIAR-RAW. On average, each claim in RAWFC is associated with 21.0 retrieved reports, whereas LIAR-RAW only has an average of 12.3 reports per claim. And the average number of sentences per report in RAWFC is higher than that in LIAR-RAW. This suggests that RAWFC provides a richer set of contextual information for fake news detection, potentially leading to more robust detection and explanations. 

\paragraph{Training Setups. }
We use the ``gpt-3.5-turbo-0125'' \cite{chatgpt} developed by OpenAI as our LLM backbone for each frozen LLM component when there is no special declaration. 
Although our framework can be easily adapted to more recent LLMs, here we choose to conduct our experiments on a widely recognized earlier model. This choice allows us to fairly demonstrate the fundamental capability of our framework rather than relying on the improvements from LLM updates. 
In the competing explanation generation module (\S \ref{subsec:explanation_gen}), we set $k = 5$ to retrieve top-5 candidate evidences for each sub-claim. For the LLM encoder used in the defense-like inference module (\S \ref{subsec:glm_inference}), we initialize it with Flan-T5-large \cite{Flan-T5} developed by Google. To fine-tune it, we use a mini-batch Stochastic Gradient Descent (SGD) to minimize the loss functions, with Adam optimizer, 10\% warm-up, and a linear decay of the learning rate. The learning rate used to fine-tune the LLM encoder is $7e^{-5}$ for RAWFC and $5e^{-5}$ for LIAR-RAW, the batch size is 8 for RAWFC and 4 for LIAR-RAW, and the training epoch is 5 for both datasets. For the temperature of LLM in our framework, we set it to 0.8, allowing the LLM to flexibly apply its internal knowledge for detection and explanation generation. We set the temperature to 0 during the evaluation of generated explanations. All experiments are conducted on a single A40 GPU.

\subsection{Veracity Prediction Performance} \label{subsec:eval_veracity}

\paragraph{Baselines. }
To compare the effectiveness of the proposed G-Defense in fake news detection, we evaluate its performance against the following baseline models, which are divided into three categories, i.e., traditional non-LLM-based approaches (1-6), LLM-based approaches (7-12), and the variants of G-Defense (13-16):
\begin{enumerate}
    \item \textbf{dEFEND} \cite{dEFEND_KDD19} introduces a GNN-based sentence-comment co-attention network for veracity prediction with explanations to jointly model sentences from news article and user comments for veracity prediction. 
    \item \textbf{SBERT-FC} \cite{EXTABS_EMNLP20_Kotonya} employs Sentence-BERT \cite{sentence-bert_Reimers_EMNLP19} to encode contextualized representations for candidate evidences and news claim. Next, it ranks evidences based on their cosine similarity with the claim and selects the top-$k$ most important sentences for veracity prediction.  
    \item \textbf{GenFE} \cite{GenFE_ACL20} employs DistilBERT \cite{distilbert} to encode the simple concatenation of a claim and its debunked journalism for veracity prediction.  
    \item \textbf{GenFE-MT} \cite{GenFE_ACL20} enhances GenFE by further introducing a multi-task setup. It jointly predicts salient sentences in news articles and the veracity of the claim.
    \item \textbf{CofCED} \cite{zhiwei22coling} utilizes DistilBERT to encode news claims and all sentences in raw reports. Then it introduces a coarse-to-fine cascaded evidence-distillation framework by focusing on key evidence to detect fake news. 
    \item \textbf{MUSER} \cite{Muser} employs Sentence-BERT to perform multi-step retrieval and build associations among pieces of evidence required for news verification. Then the semantic connection between the news claim and the evidence is analyzed with BERT for detection.
    \item \textbf{LLaMA2$_{\text{claim}}$} \cite{llama2} prompts LLaMA2-7b with the news claim to directly generate a veracity prediction and corresponding explanation.
    \item \textbf{GPT-3.5$_{\text{claim}}$} \cite{chatgpt} is similar to LLaMA2$_{\text{claim}}$, but replaces LLaMA2 with GPT-3.5. 
    \item \textbf{GPT-3.5$_{\text{full}}$} \cite{chatgpt} prompts GPT-3.5 with both the claim and all retrieved relevant reports. The absence of LLaMA2$_{\text{full}}$ is that the 7b model struggles to produce consistent output after processing such lengthy inputs. 
    \item \textbf{FactLLaMA} \cite{factllama} leverages the LORA tuning \cite{lora_ICLR22} to supervised fine-tunes a LLaMA2-7b with the claim. 
    \item \textbf{FactLLaMA$_{\text{know}}$} \cite{factllama}, compared with FactLLaMA, retrieves external evidence from search engines and feeds the LLaMA2-7b with the claim and the filtered relevant evidence.
    \item \textbf{L-Denfense$_{\text{LLaMA2}}$} and \textbf{L-Denfense$_{\text{GPT-3.5}}$} \cite{wang_www24} leverage a defense-like inference strategy on the news claim to perform classification by fine-tuning a RoBERTa model \cite{RoBERTa}, while employing LLaMA2 \cite{llama2} and GPT-3.5 \cite{chatgpt} as explanation generators respectively. 
    \item \textbf{G-Defense$_{\text{hyper}}$} replaces the dependency-based claim-centered graph with a hypergraph structure, where sub-claims are grouped into topic-centered hyperedges rather than connected through pairwise logical dependencies. The prompt $I^h$ used to instruct the LLM to generate hyperedges is provided in Appendix \ref{app:prompt}.
    \item \textbf{G-Defense$_{\text{background}}$} augments each sub-claim with additional background context. For every sub-claim, we retrieve 20 related sentences and prompt the LLM to summarize them into a concise background description. This background information is then supplied to the model to help it better understand the sub-claim and assess its veracity. The prompt $I^b$ used to instruct the LLM is provided in Appendix \ref{app:prompt}.
    \item \textbf{G-Defense$_{\text{decomp+}}$} enhances the claim decomposition stage by replacing the original simple prompt with a more structured, domain-informed version. The revised prompt incorporates decomposition strategies, news-analysis cues, dependency cues, and fact-checking dimensions. The full prompt $I^{v+}$ is provided in Appendix \ref{app:prompt}. 
    \item \textbf{G-Defense$_{\text{LLaMA3.1}}$} substitutes the GPT-based backbone with the open-source LLaMA-3.1-8B model, allowing us to explore the framework’s adaptability to different LLM backbones. 
\end{enumerate}

\begin{table}[t] 
\caption{Veracity prediction results on RAWFC and LIAR-RAW. $\dagger$Resulting numbers are reported by \citet{zhiwei22coling}, the results of FactLLaMA and L-Defense are taken from the original paper, and the results of MUSER are reproduced by us. Bold numbers indicate the best performance, and underlined numbers indicate the second-best performance. Results for G-Defense and its variants are averaged over three random runs.}
	\centering
    \renewcommand{\arraystretch}{1.2}
	\begin{tabular}{l|ccc|ccc}    \hline
	        & \multicolumn{3}{c|}{\textbf{RAWFC}}           & \multicolumn{3}{c}{\textbf{LIAR-RAW}}    \\ \cline{2-4} \cline{5-7} 
	            & \textbf{P}   & \textbf{R}   & \textbf{macF1}     & \textbf{P}   & \textbf{R}   & \textbf{macF1}  \\ \hline \hline\hline
             
		\multicolumn{7}{l}{\textit{Traditional approach}}\\ \hline
            dEFEND \cite{dEFEND_KDD19}$\dagger$    & 44.93   & 43.26   & 44.07          & 23.09   & 18.56   & 17.51 \\
            SBERT-FC \cite{EXTABS_EMNLP20_Kotonya}$\dagger$  & 51.06 & 45.92 & 45.51    & 24.09   & 22.07   & 22.19 \\
            GenFE \cite{GenFE_ACL20}$\dagger$      & 44.29   & 44.74   & 44.43          & 28.01   & 26.16   & 26.49 \\
            GenFE-MT \cite{GenFE_ACL20}$\dagger$   & 45.64   & 45.27   & 45.08          & 18.55   & 19.90   & 15.15 \\
            MUSER \cite{Muser}  & 47.47 & 47.50 & 47.48     & 27.80 & 26.58 & 26.52 \\
    		CofCED \cite{zhiwei22coling}$\dagger$  & 52.99   & 50.99   & 51.07          & 29.48   & 29.55   & 28.93 \\  
		
		\hline  \hline \hline
  
		\multicolumn{7}{l}{\textit{LLM-based approach}}\\ \hline
		LLaMA2$_{\text{claim}}$ \cite{llama}        & 37.30   & 38.03   & 36.77         & 17.11    & 17.37   & 15.14       \\
        GPT-3.5$_{\text{full}}$ \cite{chatgpt}         & 39.48   & 45.07   & 39.31         & 29.64    & 23.57   & 21.90  \\
        GPT-3.5$_{\text{claim}}$ \cite{chatgpt}      & 47.72   & 48.62   & 44.43         & 25.41    & 27.33   & 25.11       \\  \hdashline[1pt/1pt]
          
        FactLLaMA   \cite{factllama}      & 53.76   & 54.00   & 53.76         & 32.32    & 31.57    & 29.98 \\
        FactLLaMA$_{\text{know}}$  \cite{factllama}      & 56.11   & 55.50   & 55.65         & 32.46    & 32.05    & 30.44       \\ 
        
        L-Denfense$_{\text{LLaMA2}}$ \cite{wang_www24}    & 60.95     & 60.00     & 60.12     & 31.63     & 31.71     & 31.40 \\
        L-Denfense$_{\text{GPT-3.5}}$ \cite{wang_www24}   & 61.72     & 61.01    & 61.20     & 30.55     & 32.20     & 30.53 \\ \hline \hline \hline
        
        \multicolumn{7}{l}{\textit{Ours}}\\ \hline

        G-Defense    & 64.99\tiny{$\pm$0.78}     & \underline{64.52}\tiny{$\pm$1.02}   & 64.31\tiny{$\pm$1.20}           & \underline{33.09}\tiny{$\pm$1.01}        & \underline{31.55}\tiny{$\pm$0.84}     & \underline{31.55}\tiny{$\pm$0.51} \\  \hdashline[1pt/1pt]

        G-Defense$_\text{hyper}$             & 64.62\tiny{$\pm$1.79}    & 63.02\tiny{$\pm$1.31}   & 63.44\tiny{$\pm$1.43}       & 31.21\tiny{$\pm$0.59}   & 29.53\tiny{$\pm$0.64}   & 29.75\tiny{$\pm$0.57} \\
        G-Defense$_\text{background}$        & \textbf{66.29}\tiny{$\pm$1.63}    & \textbf{65.49}\tiny{$\pm$0.98}   & \textbf{65.50}\tiny{$\pm$1.08}       & 31.50\tiny{$\pm$0.89}   & 31.28\tiny{$\pm$0.62}   & 30.89\tiny{$\pm$0.50}    \\
        G-Defense$_\text{decomp+}$           & \underline{65.63}\tiny{$\pm$2.45}    & 64.33\tiny{$\pm$1.88}   & \underline{64.34}\tiny{$\pm$2.04}       & 32.03\tiny{$\pm$0.24}   & 30.23\tiny{$\pm$0.47}   & 30.36\tiny{$\pm$0.78}  \\

        G-Defense$_{\text{LLaMA3.1}}$    & 60.97\tiny{$\pm$0.84}     & 61.23\tiny{$\pm$0.27}    & 60.76\tiny{$\pm$0.76}              & \textbf{34.17}\tiny{$\pm$0.67}       & \textbf{32.37}\tiny{$\pm$0.24}     & \textbf{32.49}\tiny{$\pm$0.49} \\
        
		 \hline
	\end{tabular}
	\label{tab:veracity_results}
\end{table}

\paragraph{Metrics.}
Following previous works \cite{zhiwei22coling,factllama,wang_www24}, we use macro-average precision (P), macro-average recall (R) and macro-average F1 (macF1). They are calculated as: $\text{P} = \frac{1}{L} \sum_{l=1}^{L} \text{P}_l$, $\text{R} = \frac{1}{L} \sum_{l=1}^{L} \text{R}_l$,  $\text{macF1} = \frac{1}{L} \sum_{l=1}^{L} \frac{2 \cdot \text{P}_l \cdot \text{R}_l}{\text{P}_l + \text{R}_l}$,  
where $L$ is the total number of classes, and $\text{P}_l$ and $\text{R}_l$ denote the precision and recall for class $l$ respectively.

Table \ref{tab:veracity_results} presents the veracity prediction results of our proposed method and competitive baselines evaluated on the two benchmarks. 
We observe that our proposed G-Defense or its variants demonstrate a significant improvement over L-Defense and achieves state-of-the-art performance on all metrics. 

\begin{figure*}[t]
    \centering
    \includegraphics[width=0.45\textwidth]{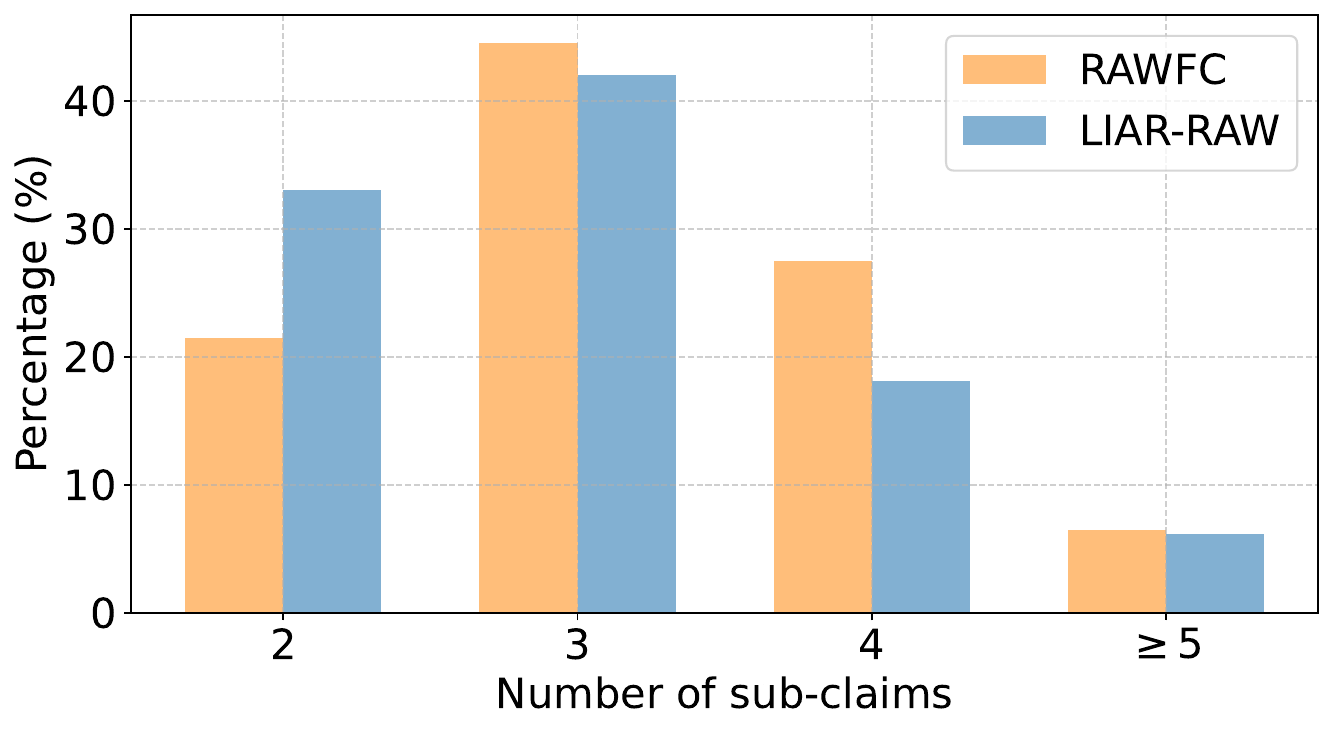}
	\caption{Distribution of claims by the number of decomposed sub-claims in the test sets of RAWFC and LIAR-RAW. 
	}
	\label{fig:sub_claims_distribution} 
\end{figure*}

\paragraph{The introduction of LLMs has significantly enhanced the performance of fake news detection. }
Among the first three LLM-based approaches without any tuning, GPT-3.5$_{\text{claim}}$ achieves the best results. 
With the exception of CofCED which uses externally retrieved reports in a coarse-to-fine evidence distillation framework, most traditional approaches underperform compared to GPT-3.5$_{\text{claim}}$. Notably, GPT-3.5$_{\text{claim}}$ operates in a zero-shot setting, making veracity predictions based solely on the claim itself without incorporating any external data. On the one hand, it suggests strong generalization capabilities and great potential of LLMs in fake news detection due to their extensive pre-training on a large corpus. On the other hand, its competitive performance may because GPT-3.5 has encountered similar claims during pretraining. 
In contrast, LLaMA2$_{\text{claim}}$ significantly underperforms, likely due to its smaller model size and pretraining on a smaller corpus compared to GPT-3.5. And GPT-3.5$_{\text{full}}$ which conducts prediction with external retrieved reports performs worse than GPT-3.5$_{\text{claim}}$ in most metrics, indicating that providing a large volume of retrieved reports without any processing may introduce noise or biases, negatively impacting prediction accuracy. 
Apart from the three LLM-based methods under a zero-shot setting, all fine-tuned models combined with LLMs outperform all traditional approaches. 
FactLLaMA and FactLLaMA$_{\text{know}}$ simply fine-tune LLaMA2 with LoRA without any special design, while they show much better results than LLaMA2$_{\text{claim}}$. This suggests that fine-tuning can significantly enhance the capability of LLMs in fake news detection, enabling them to better handle complex reasoning tasks. Furthermore, it highlights the limitations of traditional methods, which struggle to compete with fine-tuned LLMs due to their lack of deep language understanding. 
As for L-Defense, it fine-tunes a small language model but incorporates the LLM's powerful reasoning and generation ability into its specifically designed framework, leading to a better result than fine-tuned LLaMA2. This highlights the importance of designing well-structured models to effectively leverage the powerful capabilities of LLMs.

\paragraph{G-Defense consistently outperforms prior methods across both benchmarks. }
Compared with L-Defense, our proposed G-Defense employs defense-like inference on a claim-centered graph using LLM, achieving excellent performance in veracity prediction on both benchmarks. 
In addition, to further validate the statistical significance of the improvements, we conduct paired t-tests over five random seeds on macF1 between G-Defense and the strongest baseline, L-Defense. The improvements are highly significant on both datasets, with $p=3\times10^{-4}$ on RAWFC and $p=9\times10^{-5}$ on LIAR-RAW. The results suggest the effectiveness of claim decomposition and graph construction in improving veracity prediction, which will be further discussed in \S \ref{subsec:ablation}. By structuring the claim, sub-claims, and competing explanations into a graph, G-Defense enhances its inference process, allowing for a more comprehensive and more accurate classification. 
The improvement is especially significant on RAWFC. Compared to the previous SOTA method, L-Defense$_{\text{GPT-3.5}}$, it improves performance by at least 3.1\% across all evaluation metrics. On LIAR-RAW, while G-Defense achieves slightly lower recall than L-Defense$_{\text{GPT-3.5}}$, it shows improvements of 2.54\% in precision and 1.02\% in macF1. 
These performance differences can be attributed to the inherent differences in data characteristics between RAWFC and LIAR-RAW.
First, as shown in Table \ref{tb:datasets}, RAWFC provides more reports per claim and more sentences per report compared to LIAR-RAW. This richer evidence context benefits explanation generation for each sub-claim and improves veracity prediction.
Second, claims in RAWFC are generally more complex than those in LIAR-RAW. We compare the average token lengths of claims and find that RAWFC claims are longer on average (28.1 tokens) than those in LIAR-RAW (23.8 tokens). Furthermore, we present the distribution of claims by the number of decomposed sub-claims in the test sets of both datasets in Figure \ref{fig:sub_claims_distribution}. It shows that claims in RAWFC are generally decomposed into more sub-claims than those in LIAR-RAW. These observations indicate the claims in RAWFC are more complex. The significant improvements on RAWFC highlight the effectiveness of our method in handling more complex claims.

\paragraph{The framework of G-Defense shows great potential in enhancing the performance of fake news detection. }
To explore the potential of our proposed G-Defense framework, we further explore four variants of it. 
G-Defense$_{\text{hyper}}$ performs consistently worse than G-Defense on both datasets, indicating that grouping sub-claims solely by topical similarity is insufficient for veracity prediction. Fine-grained logical dependencies provide clearer reasoning paths, enabling more accurate aggregation of evidence. When dependencies do not exist, sub-claims typically correspond to different facets of the main claim and contribute independently to the final prediction, which the original design naturally accommodates. 
G-Defense$_{\text{background}}$ shows significant gains on RAWFC, improving macF1 from 64.31 to 65.50, but leads to a slight decrease on LIAR-RAW. RAWFC offers longer claims and richer reports, making background consolidation more beneficial for guiding sub-claim detection, whereas LIAR-RAW’s shorter, self-contained claims leave less room for improvement. This result suggests that incorporating additional, informative context remains a promising direction for further enhancement.
G-Defense$_{\text{decomp+}}$ uses the enhanced prompt for claim decomposition and shows slight gains in precision and macF1 on RAWFC, indicating that the additional guidance helps generate clearer sub-claims for fact-rich and complex claims. On LIAR-RAW, however, we observe a slight performance drop. Claims in LIAR-RAW are generally shorter and contain fewer explicit factual components, leaving limited room for additional decomposition cues to introduce meaningful structure. In such cases, the stronger prompt may lead to unnecessary over-splitting, which may slightly disturb downstream inference. These observations indicate that the effectiveness of enhanced decomposition guidance depends on the dataset and the inherent structure of the claims. 
Finally, G-Defense$_{\text{LLaMA3.1}}$ surpasses G-Defense on LIAR-RAW on all metrics, and shows small improvements on RAWFC compared with L-Defense{$_\text{LLaMA2}$}. This demonstrates that G-Defense is not tied to a specific LLM backbone and can effectively leverage models with different parameter scales. 
Overall, these variants collectively show that G-Defense is a flexible and extensible framework whose performance can be further improved through enriched context, enhanced decomposition prompts, or newer LLM backbones, allowing users to select the configuration best suited to their application needs.

\subsection{Ablation Study} \label{subsec:ablation}
To evaluate the contribution of each component, we conduct an extensive ablation study for G-Defense on veracity prediction by removing or replacing the key component: 
\begin{enumerate}
    \item ``\textbf{w/o sub-claims}'': It does not perform claim decomposition or graph construction. In the competing explanation generation module (\S \ref{subsec:explanation_gen}), it retrieves relevant evidences for the entire claim and directly generates competing explanations for the entire claim. 
    \item ``\textbf{w/o edges}'': This variant removes the edge generation process in \S\ref{subsec:graph_construction} which creates dependency edges between sub-claims. During the defense-like inference (\S\ref{subsec:glm_inference}), it removes the graph structure description $\rmH_{\tilde \gG}$ in Eq.(\ref{eq:graph_encoding}). 
    \item ``\textbf{w/o evidence}'': It removes the evidence retrieval step in \S\ref{subsec:explanation_gen}. As a result, the LLM generates veracity-oriented explanations for sub-claims using only its own knowledge rather than retrieved external information.
    \item ``\textbf{w/o competing}'': This variant removes the prior veracity label for sub-claims during explanation generation with LLM in \S\ref{subsec:explanation_gen}. The LLM is then instructed to generate only one explanation based on its analysis rather than considering competing explanations.
    \item ``\textbf{w/o inference training}'': It replaces the fine-tuning process of defense-like inference in \S\ref{subsec:glm_inference} with direct predictions from GPT-3.5, which allow G-Defense to conduct in a zero-shot setting.
\end{enumerate}

\begin{table}[] 
\caption{Ablation study of veracity prediction on two datasets.} 
    \centering
    \renewcommand{\arraystretch}{1.2}
    \begin{tabular}{l|ccc|ccc}
\hline
& \multicolumn{3}{c|}{\textbf{RAWFC}}           & \multicolumn{3}{c}{\textbf{LIAR-RAW}}    \\ \hline
\textbf{Method}             & \textbf{P} & \textbf{R} & \textbf{macF1} & \textbf{P} & \textbf{R} & \textbf{macF1} \\ \hline 
G-Defense                   & 65.89     & 65.54     & 65.65   & 32.94     & 32.09     & 32.09   \\ \hline

~~w/o sub-claims            &56.29      &56.02      & 56.10      &30.94      &28.13  &28.52  \\
~~w/o edges                  & 64.12      & 63.49      & 63.40   & 31.39     & 30.52     & 30.60 \\
~~w/o evidence              & 56.68      & 55.98      & 55.76   & 30.12     & 29.44     & 29.30 \\
~~w/o competing             & 57.24      & 55.96      & 55.94   & 30.42     & 29.53     & 29.46 \\
~~w/o inference training    & 54.71      & 51.48      & 47.68    & 28.24     & 25.13     & 19.45 \\   \hline
\end{tabular}
    
    \label{tab:ablation}
\end{table}

As shown in Table \ref{tab:ablation}, the ablative variants suffer different degrees of performance degradation, indicating the effectiveness of each component in our proposed G-Defense for fake news detection. 
The ``w/o sub-claims'' variant shows a significant decline because the model loses the fine-grained factual units and their corresponding evidence needed for careful reasoning. Without decomposition, the LLM must judge the entire claim at once, which makes it difficult to identify the key factual conditions that determine truthfulness.
Removing edges causes a smaller drop, indicating that while sub-claims provide most of the useful information, the dependency edges help connect them into a clear reasoning flow. Without these dependency relations, the model cannot easily understand how different aspects of the claim relate to each other.
In the ``w/o evidence'' variant, performance decreases because the model can no longer rely on externally retrieved raw reports. LLMs often lack recent or complete knowledge especially for breaking news, so removing retrieved evidence forces the model to depend only on what it already knows, which limits accuracy. This shows that external evidence is important for grounding the reasoning process. 
This result underscores the necessity of incorporating RAG in fake news detection to provide informative evidence and improve model performance. 
The ``w/o competing'' variant performs worse because relying on a single explanation can be one-sided. The competing-explanation mechanism encourages the model to consider both supporting and opposing views, which reduces bias and helps prevent mistaken conclusions. 
Finally, the ``w/o inference training'' variant achieves the worst result among all variants, which proves the importance of fine-tuning in such a classification task, consistent with previous findings \cite{aaai24_hu,LLM_biased_su_arxiv23}. Without training, the classifier cannot learn how to integrate graph features, so it falls back to a purely zero-shot prediction strategy. Notably, even under this condition, this variant still obtains acceptable performance and outperforms many traditional approaches. This further highlights the superiority of our proposed framework.

\subsection{Impact of Claim Complexity} \label{subsec:complexity_analysis}

\begin{figure}[t]
    \centering
    \includegraphics[width=0.48\linewidth]{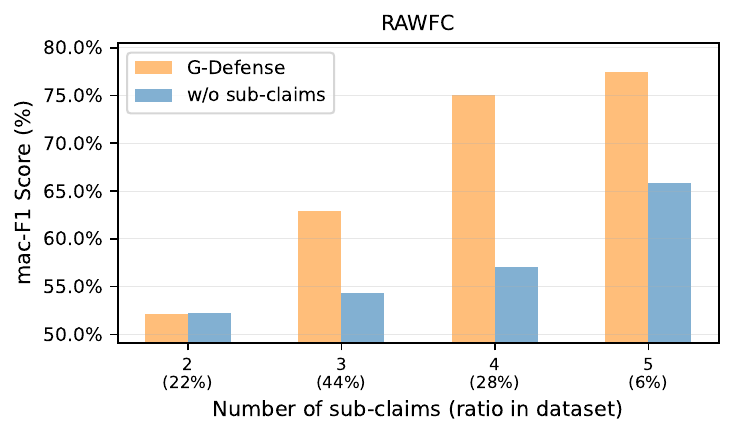}
    \hfill
    \includegraphics[width=0.48\linewidth]{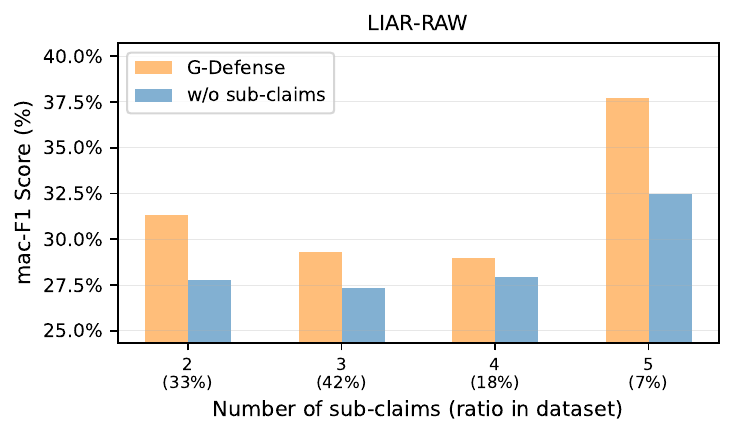}
    \caption{MacF1 scores of G-Defense and its w/o-sub-claims variant grouped by the number of sub-claims on the RAWFC (left) and LIAR-RAW (right) test sets. The percentages under each group indicate the proportion of samples within that group.}
    \label{fig:complexity_analysis}
\end{figure} 

To examine when claim decomposition is most beneficial, we group test samples by the number of decomposed sub-claims $n$. This measure reflects the factual complexity of a claim more reliably than surface length. And we compare G-Defense with its ``w/o sub-claims'' variant described in Section \ref{subsec:ablation}.
The results are shown in Figure \ref{fig:complexity_analysis}. When a claim contains only two sub-claims, G-Defense performs better on LIAR-RAW but slightly worse on RAWFC. A possible reason is that decomposing very simple claims is unnecessary and may introduce minor noise. 
As complexity increases ($n \ge  3$), the pattern changes consistently. G-Defense outperforms the ``w/o sub-claims'' across both datasets, and the performance gap grows as the number of sub-claims becomes larger. Claims with more sub-claims usually involve multiple aspects, conditions, or factual components. Predicting them as a single long sentence can be unstable, because different pieces of information may interfere with each other. Claim decomposition provides fine-grained reasoning units and allows the graph structure to integrate evidence more reliably. 
More than half of all test samples fall into the $n \ge 3$ range. This indicates that the cases where decomposition brings meaningful benefits cover a large portion of real-world claims. Although it may cause slight performance drops for the simplest cases, we apply the same decomposition procedure to all claims regardless of their complexity. Claim complexity is difficult to judge automatically, and even short claims may imply multiple facts. Using a unified detection pipeline also ensures consistent graph-structured explanations. 
Overall, this analysis shows that claim decomposition is particularly effective for claims involving more aspects or factual elements.

\vspace{6pt}
\subsection{Explanation Evaluations} \label{subsec:eval_explanation}

\paragraph{Baselines. } 

Based on the veracity prediction results in Table \ref{tab:veracity_results}, we introduce the following baseline models for explanation evaluations: 
\begin{enumerate}
    \item \textbf{Oracle}, which offers detailed justifications for the claim's classification according to its true veracity label. It achieves this by supplying GPT-3.5 with both the claim and its actual veracity label.
    \item \textbf{CofCED} \cite{zhiwei22coling}, recognized as the best model among traditional approaches. 
    \item \textbf{GPT-3.5$_{\text{claim}}$}, which exhibits the best veracity prediction performance among LLM-based approaches without any training, providing explanations according to its veracity predictions. 
    \item \textbf{GPT-3.5$_{\text{full}}$}, which is provided with retrieved raw reports for comparison with \textbf{GPT-3.5$_{\text{claim}}$}, providing explanations according to its veracity predictions. 
    \item \textbf{L-Defense$_\text{LLaMA2}$} and \textbf{L-Defense$_{\text{GPT-3.5}}$} \cite{wang_www24}, which is the previous SOTA model. Since its generated explanations are limited to three categories: \textit{true}, \textit{half-true}, and \textit{false}, we map the six labels in the LIAR-RAW dataset to these three categories before getting explanations. Specifically, \textit{pants-fire}, \textit{false}, and \textit{barely-true} are grouped under the \textit{false} category. The \textit{half-true} category remains unchanged, while \textit{mostly-true} and \textit{true} are merged into the \textit{true} category.
\end{enumerate}

\paragraph{Metrics. } 
Traditional automated evaluation metrics like ROUGE are often insufficient for assessing the quality of output generated by LLMs \cite{survey_LLM_eval}, as they primarily measure surface-level n-gram overlaps and fail to capture semantic accuracy and reasoning quality. 
Recently, \citet{llm-for-eval_Chen_arxiv23} has demonstrated that GPT performs well in evaluating text quality from multiple perspectives, even without reference texts. Additionally, studies have shown that LLM-based evaluation closely aligns with expert human assessments \cite{LLM-eval_ACL23_Chiang,exp-eval-figure_WWW23_Huang}.
Therefore, we utilize GPT-3.5 to evaluate the quality of explanations based on four widely adopted human evaluation metrics \cite{zhiwei22coling,exp-eval-metric_IJCAI23_Wang}: \textit{misleadingness}, \textit{informativeness}, \textit{soundness}, and \textit{readability}. Each metric is rated on a 5-point Likert scale, where 1 represents the lowest score and 5 the highest, except for misleadingness. The definitions of these metrics are as follows:
\begin{itemize}
    \item[---] \textbf{Misleadingness (M)} measures whether the explanation aligns with the true veracity label of a claim. It is rated from 1 (not misleading) to 5 (highly misleading);
    \item[---] \textbf{Informativeness (I)} measures the extent to which the explanation provides new insights, such as background details and additional context. The scale ranges from 1 (not informative) to 5 (highly informative); 
    \item[---] \textbf{Soundness (S)} describes whether the explanation seems valid and logical, with a rating scale ranging from 1 (not sound) to 5 (very sound);
    \item[---] \textbf{Readability (R)} assesses whether the explanation follows proper grammar and structural rules, and whether the sentences in the explanation fit together and are easy to follow, with a rating scale ranging from 1 (poor) to 5 (excellent). 
\end{itemize}
To further validate the reliability of LLM-based evaluation, we introduce an automated metric called \textbf{Discrepancy (D)}. Unlike misleadingness, which relies on LLM assessments of explanations, discrepancy is directly computed based on the consistency between the predicted and gold veracity labels without involving LLM evaluation. Specifically, it is defined as the absolute difference between the predicted and true labels.
For RAWFC, we map the three veracity labels \{\textit{false}, \textit{half}, \textit{true}\} to numerical scores \{0, 2.5, 5\}, and for LIAR-RAW, the six fine-grained labels are mapped to \{0, 1, 2, 3, 4, 5\} based on their annotated truthfulness degrees. This automatic metric provides an objective measure that complements the LLM-based misleadingness evaluation. The consistency between discrepancy and misleadingness can reflect the reliability of the LLM-based evaluation.

\begin{table}[t]
\caption{Evaluation results of explanation quality using a 5-Point Likert scale rating by ChatGPT on RAWFC and LIAR-RAW. For metrics D and M, a lower score ($\downarrow$) indicates better performance, while a higher score ($\uparrow$) indicates better performance for the remaining metrics. The bold numbers denote the best results in addition to Oracle while the underlined ones are better than Oracle. } 
\centering
\renewcommand{\arraystretch}{1.2}
    \begin{tabular}{l|c|cccc|c|cccc}
    \hline
                      & \multicolumn{5}{c|}{\textbf{RAWFC}}          & \multicolumn{5}{c}{\textbf{LIAR-RAW}}     \\ \cline{2-11} 
                      & \textbf{D}$\downarrow$ & \textbf{M}$\downarrow $    & \textbf{I}$\uparrow$     & \textbf{S}$\uparrow$     & \textbf{R}$\uparrow$                & \textbf{D}$\downarrow$   & \textbf{M}$\downarrow$     & \textbf{I}$\uparrow$    & \textbf{S}$\uparrow$    & \textbf{R}$\uparrow$    \\ \hline
    Oracle            &  -            & 1.52  & 4.46  & 4.73   & 4.72      & -      & 1.85 & 4.44 & 4.60  & 4.69 \\ \hline
    CofCED \cite{zhiwei22coling}   & 1.53        & 2.74    & 2.89          & 1.93   & 2.46    & 1.33 & 3.64 & 1.75 & 1.76  & 1.59 \\
    
    GPT-3.5$_\text{claim}$         & 1.70 & 1.97   & 4.00  & 4.44   & 4.68       & 1.39   & 2.27 & 3.93 & 4.29  & 4.50 \\ 
    GPT-3.5$_\text{full}$          & 1.81  & 2.07  & 4.44 & 4.62  & 4.69        & 1.39   & 2.29 & 3.71 & 4.04  & 3.99 \\
    L-Defense$_\text{LLaMA2}$ \cite{wang_www24} & 1.30  & 1.95 & 4.44 & 4.67   & 4.62         & 1.36   & 2.20 & 4.39 & \underline{\textbf{4.64}}  & 4.63 \\
    L-Defense$_\text{GPT-3.5}$ \cite{wang_www24}  & 1.30  & 1.91  & 4.17  & 4.41   & 4.49     & 1.31   & 2.06 & 4.12 & 4.28  & 4.47 \\ \hline
    G-Defense &\textbf{1.00} &\textbf{1.87} &\underline{\textbf{4.60}} &\textbf{4.72} &\textbf{4.70} &\textbf{1.26} &\textbf{2.02} &\textbf{4.42} &4.38 &\textbf{4.65}\\ \hline
    \end{tabular}
    \label{tab:gpt_explanation_eval}
\end{table}

We list the results of the LLM-based explanation evaluation in Table \ref{tab:gpt_explanation_eval}. 
The Oracle, which is prompted with the ground-truth labels, outperforms all other methods across nearly all metrics and can be regarded as an upper bound for explanation quality. In comparison, G-Defense achieves either superior or comparable performance,  and consistently outperforms all other baselines across most metrics, indicating its strong capability to generate reasonable and reliable explanations.
In particular, it obtains the lowest discrepancy (D) and lowest misleadingness (M) scores on both datasets, suggesting that the explanations produced by G-Defense are highly consistent with the ground-truth veracity labels. This is especially important, as a key goal of explainable fake news detection is to assist the public in identifying the truth, rather than misleading users through incorrect or biased reasoning. 
Moreover, the strong correlation between the objective discrepancy scores and the LLM-assessed misleadingness scores further validates the reliability of our evaluation methodology.

As for other baselines, as the best traditional approach, CofCED achieves the worst results on all LLM-evaluated metrics. 
Since its explanations are discrete sentences extracted from reports, which often contain redundant information and lack coherence, making them less human-friendly. This incoherent explanation leads to a high misleadingness score, though the objective discrepancy score is much better. And the other three worst metrics also underscore the importance of coherence in generating a streamlined and understandable explanation. 
In terms of two baselines based on GPT-3.5, GPT-3.5$_{\text{full}}$ generally outperforms GPT-3.5$_{\text{claim}}$ on RAWFC while underperforming it on LIAR-RAW. 
This discrepancy may result from the larger number of candidate evidences in RAWFC, which provides more information to support explanation generation.

Compared with both versions of L-Defense, our proposed G-Defense demonstrates consistent improvements on almost all metrics. 
By leveraging more fine-grained sub-claims, along with their corresponding veracity assessments and explanations, G-Defense is able to deliver richer and more informative explanations of the claim's veracity. This is reflected in the higher information scores (I) compared to L-Defense. 
And notably, it even outperforms the Oracle baseline on RAWFC. 
Moreover, this fine-grained explanation strategy contributes to improvements not only in informativeness but also in soundness (S) and readability (R). The decomposition into sub-claims allows the model to analyze specific aspects of the overall claim, resulting in explanations that are logically more robust and easier for readers to follow. 
Overall, the results suggest that breaking down complex claims into smaller, verifiable units is not only beneficial for model prediction, but also crucial for generating clear, trustworthy, and human-friendly explanations, which is a key requirement for real-world fake news detection systems.

\begin{figure}[t]
    \centering
    \includegraphics[width=0.9\textwidth]{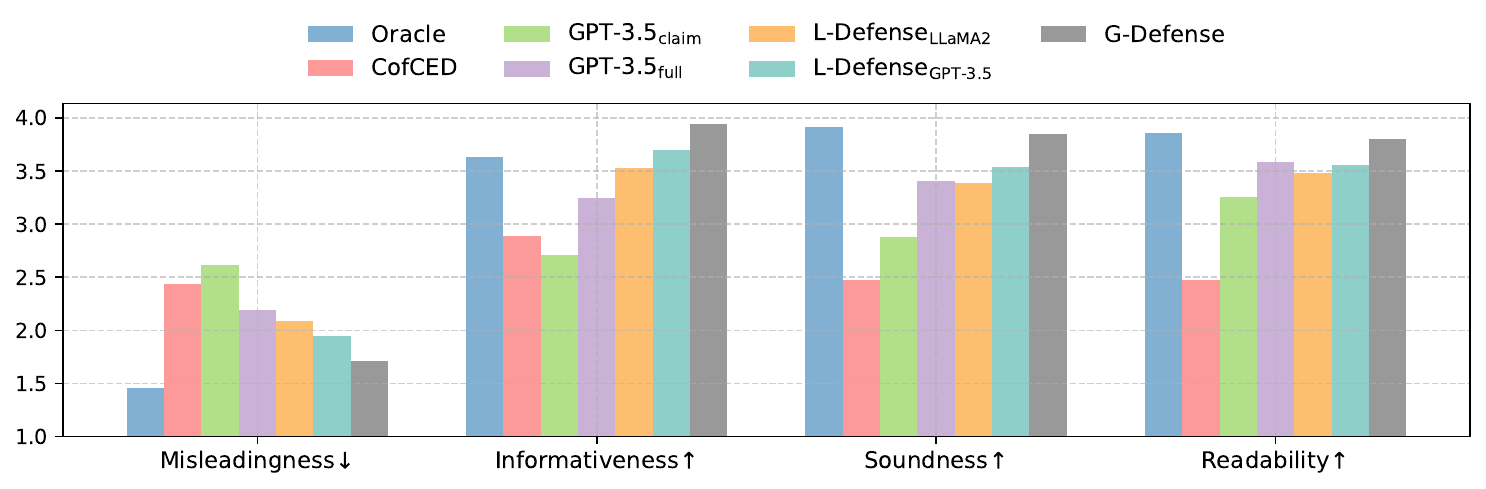}
	\caption{Explanation evaluation results using a 5-Point Likert scale rating by five human annotators on 30 randomly sampled claims from the test set of RAWFC. Scores from five annotators were averaged. }
	\label{fig:human_eval_metric} 
\end{figure}

\paragraph{Human Evaluation. }

While LLM-based evaluation provides a scalable and cost-efficient way to assess the quality of generated explanations, it cannot fully reflect how humans judge these explanations. To complement the automatic assessment, we further conduct a human evaluation. Specifically, we randomly sample 30 claims from the test set of RAWFC and engage five English-speaking adults as annotators to assess the quality and helpfulness of the explanations. For the explanations generated by G-Defense, we provide annotators with both the explanation graph and the corresponding textual description. Additional reliability analyses of the human evaluation are provided in Appendix \ref{app:humaneval_reliability}.

We present the results of human evaluation in Figure \ref{fig:human_eval_metric}. Compared with the LLM-based evaluation results on RAWFC shown in Table \ref{tab:gpt_explanation_eval}, the human ratings exhibit a generally consistent trend across all methods, further validating the reliability of our LLM-based evaluation. 
CofCED ranks the lowest on soundness and readability, likely due to the discrete and fragmented nature of its explanations, which hampers human understanding. 
Without any external relevant reports provided, GPT-3.5$_{\text{claim}}$ performs the worst in terms of informativeness. And its performance is consistently inferior to GPT-3.5${_\text{full}}$, highlighting the importance of RAG in fake news detection. 
In contrast, both L-Defense and G-Defense generally achieve higher ratings, suggesting that the defense-like inference strategy contributes to generating higher-quality explanations.
Notably, G-Defense outperforms all other methods across all metrics except the Oracle, further demonstrating the effectiveness of fine-grained explanations enabled by claim-centered graph construction.

\begin{figure}[t]
    \centering
    \includegraphics[width=0.7\textwidth]{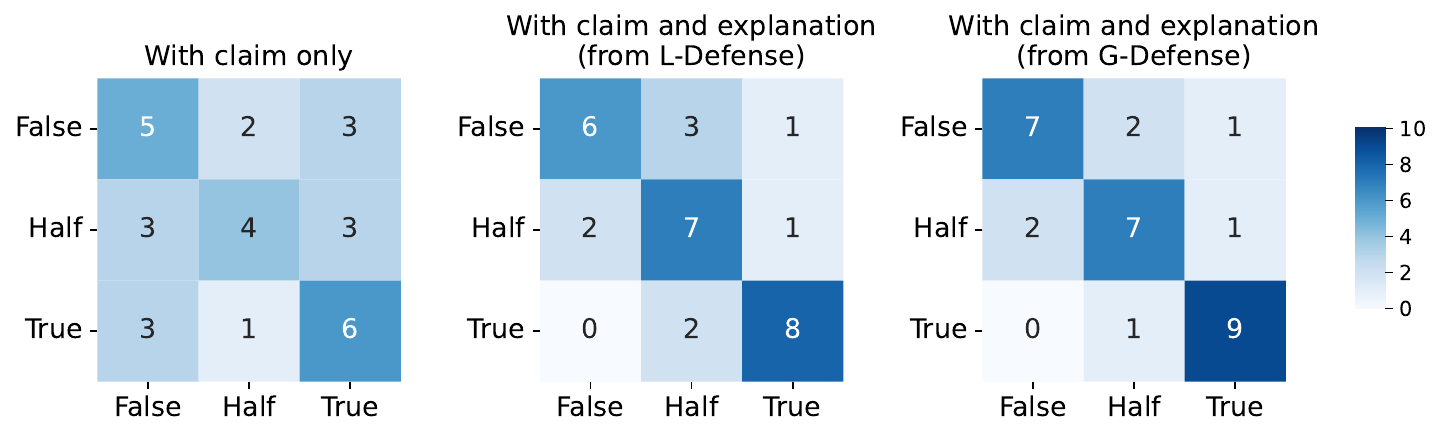}
	\caption{Confusion matrices of judgment results from five annotators on 30 randomly sampled claims. The three results are collected under three different settings: annotators are provided with only the claim, with the claim and explanations generated by L-Defense, or with the claim and explanations generated by G-Defense. The results from five annotators were averaged and rounded for visualization. }
	\label{fig:human_eval_confusion} 
\end{figure}

In addition to assessing the quality of explanations, we ask the annotators to judge the veracity of claims with and without explanation, and present the results in the form of confusion matrices in Figure \ref{fig:human_eval_confusion}. 
This human-centered evaluation directly examines the usefulness and impact of the generated explanations on real users' decision-making. We observe that providing explanations remarkably reduces judgment errors and helps annotators better identify the truth, confirming the practical utility and trustworthiness of explanations in alleviating the propagation of fake news. When comparing the two settings where explanations are provided, G-Defense is slightly better than L-Defense. These results demonstrate that the intuitive and fine-grained explanations produced by G-Defense enable annotators to better distinguish subtle differences between veracity levels and have a measurable positive effect on human judgments.

\begin{table}[t] \scriptsize
\caption{Case study. The claims are selected from the test set of RAWFC. We provide both competing explanations for analysis. In practice, only the explanation corresponding to the predicted label would be presented to users, and the other one (shown in gray) is included here for analysis. The gold label of the claim and the predicted labels of both the claim and its sub-claims are highlighted in color.}
\centering
\renewcommand{\arraystretch}{1.1}
\begin{tabular}{@{} m{0.15\linewidth} @{\hskip 1em} m{0.8\linewidth} @{}}
\hline

\multicolumn{2}{p{0.95\linewidth}}{
    \href{https://www.snopes.com/fact-check/dermophis-donaldtrumpi-amphibian-trump/}{claim 1} ({\color{green}True}): A newly discovered species of amphibian was named ``Dermophis donaldtrumpi'' as an unflattering reference to the U.S. President.
} \\ \hdashline[1pt/1pt]
\adjustbox{valign=t}{\includegraphics[width=\linewidth]{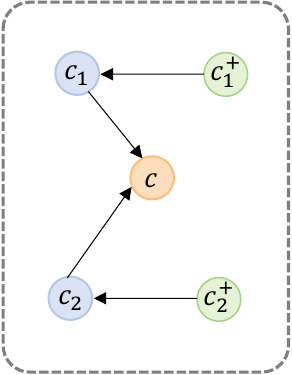}} & 
\parbox[t]{\linewidth}{
$c_1$ ({\color{green}True}): A newly discovered species of amphibian was named ``Dermophis donaldtrumpi''. \\
$c_1^+$: The newly discovered species of amphibian was named ``Dermophis donaldtrumpi'' in recognition of the US president\'s climate change denial. \\
{\color{gray}$c_1^-$: The claim is false because there is no newly discovered species of amphibian named ``Dermophis donaldtrumpi''}\\
$c_2$ ({\color{green}True}): The name ``Dermophis donaldtrumpi'' was chosen as an unflattering reference to the U.S. President. \\
$c_2^+$: The name ``Dermophis donaldtrumpi'' was chosen as an unflattering reference to the U.S. President because EnviroBuild, who paid for the naming rights, stated that they chose the name in recognition of the president\'s position on climate change.\\
{\color{gray}$c_2^-$: The name ``Dermophis donaldtrumpi'' was chosen as a reference to the U.S. President\'s climate change policy.} 
} \\
\multicolumn{2}{p{0.95\linewidth}}{
    \textbf{Summary}: The claim is reasoned as `{\color{green} true}' based on the supporting evidence provided by the true-oriented explanations for both sub-claims, which establish a direct connection between the naming of the species and the U.S. President's stance on climate change. 
} \\ \hline

\multicolumn{2}{p{0.95\linewidth}}{
    \href{https://www.snopes.com/fact-check/aoc-fdr-amendment/}{claim 2} ({\color{orange}Half}): The U.S. Congress passed the 22nd Amendment in order to ``make sure President Franklin D. Roosevelt did not get re-elected''.
} \\ \hdashline[1pt/1pt]
\adjustbox{valign=t}{\includegraphics[width=\linewidth]{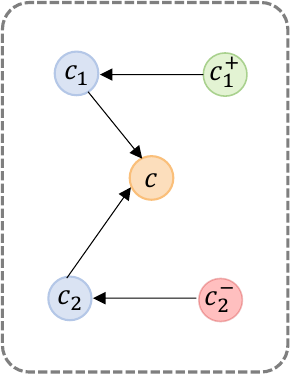}} & 
\parbox[t]{\linewidth}{
$c_1$ ({\color{green}True}): The U.S. Congress passed the 22nd Amendment. \\
$c_1^+$: The U.S. Congress passed the 22nd Amendment in 1947, which was later ratified by the states in 1951. \\
{\color{gray}$c_1^-$: The 22nd Amendment was approved by Congress in 1947, with an exception that would exclude a president in office from term limit during the ratification process.}\\
$c_2$ ({\color{red}False}): The purpose of the 22nd Amendment was to prevent President Franklin D. Roosevelt from being re-elected. \\
{\color{gray}$c_2^+$: The 22nd Amendment was not intended to prevent President Franklin D. Roosevelt from being re-elected, but rather to establish a term limit for all future presidents.}\\
$c_2^-$: The 22nd Amendment was not intended to prevent President Franklin D. Roosevelt from being re-elected, but rather to establish a term limit for all future presidents. FDR died in 1945. The amendment was passed in 1947 and ratified in 1951 — after his death. 
} \\
\multicolumn{2}{p{0.95\linewidth}}{
    \textbf{Summary}: The claim is reasoned as `{\color{orange} half}' because while it is true that the U.S. Congress passed the 22nd Amendment, the assertion that it was solely to prevent President Franklin D. Roosevelt from being re-elected is false. The actual purpose was to set a term limit for all future presidents.
} \\ \hline

\multicolumn{2}{p{0.95\linewidth}}{
    \href{https://www.snopes.com/fact-check/covid-19-infection-aspirin/}{claim 3} ({\color{red}False}): By ignoring a ban on performing COVID-19 autopsies, researchers in Italy discovered that the disease is caused by a bacteria — not a virus — and can simply be treated with aspirin. 
} \\ \hdashline[1pt/1pt]
\adjustbox{valign=t}{\includegraphics[width=\linewidth]{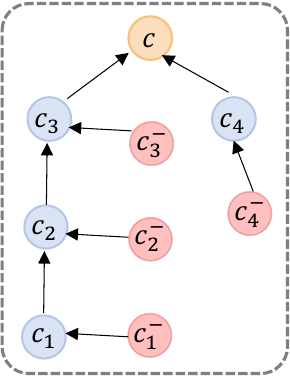}} & 
\parbox[t]{\linewidth}{
$c_1$ ({\color{red}False}): There was a ban in Italy on performing autopsies on COVID-19 victims. \\
{\color{gray} $c_1^+$: Early in the pandemic, some countries, including Italy, issued temporary restrictions or guidelines discouraging autopsies due to safety concerns.} \\
$c_1^-$: It is false because there was no absolute ban. Autopsies were conducted under strict biosafety protocols. The claim exaggerates the extent of any restriction.\\
$c_2$ ({\color{red}False}): Researchers ignored the ban and performed unauthorized autopsies. \\
{\color{gray}$c_2^+$: The claim is reasoned as true because there is evidence of researchers evaluating hospitalized COVID-19 patients to determine the risk of death from COVID-19.}\\
$c_2^-$: It is false because no evidence supports the claim that any autopsies were illegal or conducted in defiance of restrictions. \\
$c_3$ ({\color{red}False}): The autopsies revealed that COVID-19 is caused by bacteria, not a virus.\\
{\color{gray}$c_3^+$: It is true because researchers misunderstood co-infections or inflammation due to secondary bacterial infections, not the cause of COVID-19.}\\
$c_3^-$: The claim is false because multiple studies mentioned in the sentences indicate that COVID-19 is caused by a virus, not a bacteria. \\
$c_4$ ({\color{red}False}): COVID-19 can be easily treated with aspirin.\\
{\color{gray}$c_4^+$: Aspirin use may have a beneficial effect in COVID-19 patients, potentially reducing mortality.}\\
$c_4^-$: The claim is false because there is no confirmed evidence that aspirin can treat COVID-19. 
} \\
\multicolumn{2}{p{0.95\linewidth}}{
    \textbf{Summary}: This claim reasoned as `{\color{red} false}'. While there were limited restrictions on autopsies in Italy early in the pandemic, there is no evidence that researchers violated bans or discovered that COVID-19 is caused by bacteria. 
} \\ \hline

\end{tabular}

\label{tab:case_study}
\end{table}

\subsection{Case Study} \label{subsec:case_study}
To qualitatively demonstrate the effectiveness of our proposed G-Defense framework, we present three samples selected from the test set of RAWFC in Table \ref{tab:case_study}. Each case consists of a claim, its corresponding explanation graph, and the generated textual explanation. 

\paragraph{Does the claim decomposition and graph construction help veracity prediction and provide a fine-grained explanation?} 
As shown in Table \ref{tab:case_study}, all three claims are decomposed into at least two sub-claims, with each sub-claim capturing a distinct aspect of the original claim. This decomposition enables more targeted reasoning and provides more comprehensive evidence. 
For example, in claim 1, the true-oriented explanation $c_2^+$ for sub-claim $c_2$ introduces additional knowledge that EnviroBuild paid for the naming rights as a reference to the U.S. President’s stance on climate change. This reveals the potential motivation behind the naming, providing more details to support the claim as true. 
In claim 2, the two sub-claims are evaluated using different historical facts: the exact year the 22nd Amendment was passed and ratified, and the death date of Franklin D. Roosevelt. This decomposition enables the model to identify that, although the amendment itself is factual, the implication of the claim is misleading. As a result, the model assigns a ‘half’ label. 
In addition to improving veracity prediction, the explanation graphs provide an intuitive visual format that reflects the internal reasoning process. They clearly display the veracity of sub-claims, their supporting or refuting explanations, and how they influence the final decision based on the dependency relationships. This transparent reasoning path enhances explainability, and reduces the reading cost for users if they only want to know which part is false without caring about the specific details. 
These samples demonstrate that claim decomposition and graph construction are crucial for both accurate veracity prediction and intuitive explanation generation.

\paragraph{How does the defense-like inference strategy improve the performance of fake news detection?}
Our method generates a pair of competing explanations for each sub-claim. By implicitly comparing the informativeness and confidence of two explanations, the model can make more accurate inferences than relying on one-sided explanations. 
On the one hand, explanations that do not align with the true label often lack strong reasoning or supporting evidence. 
For instance, in claim 1, the false-oriented explanation $c_1^-$ for sub-claim $c_1$ is very shallow and does not contain any meaningful information, especially when compared to its true-oriented explanation $c_1^+$. 
Similarly, in claim 3, $c_4^+$ offers an unconfident explanation, whereas $c_4^-$ is much more direct and informative. 
These contrasts help the model identify which explanation is more credible, so that improving veracity prediction. 
On the other hand, even incorrect explanations can sometimes indirectly reveal the truth by exposing logical inconsistencies. 
For example, in claim 1, $c_2^-$ who aims to refute $c_2$ actually supports it. Likewise, $c_2^+$ in claim 2 proves the $c_2$ as false. And $c_2^+$ and $c_3^+$ in claim 3 fail to directly explain why the sub-claims are true, but offer reasoning that indirectly demonstrates the sub-claims are false. 
This may be because, without solid supporting evidence, LLMs struggle to fabricate convincing yet incorrect explanations.
In summary, the defense-like inference strategy reduces dependence on potentially inaccurate or biased single-perspective explanations. 
Instead, it shifts the focus to comparative reasoning, encouraging the model to evaluate the quality of competing explanations, thereby improving the performance of fake news detection. 

\vspace{20pt}
\subsection{Discussion} \label{subsec:error_propa_discussion}
Although G-Defense involves multiple stages, it is inherently robust to error propagation. 
In the graph construction module, we carefully design the prompt to encourage the generation of all relevant sub-claims. To ensure minimal fine-grained reasoning capability, we require that each claim be decomposed into at least two sub-claims. And we introduce an additional safeguard to enhance graph connectivity by explicitly adding edges between each sub-claim and the original claim. This ensures that even if the LLM fails to capture all dependencies among sub-claims and claims, the resulting graph remains connected, preventing the presence of isolated nodes that could compromise downstream reasoning. 
In the competing explanation generation module, as discussed in \S\ref{subsec:case_study}, the LLM may sometimes struggle to generate a convincing explanation supporting an incorrect veracity label due to the lack of sufficient evidence. This limitation actually helps veracity prediction, as poor-quality explanations for incorrect labels naturally guide the inference module toward more accurate reasoning. 
Moreover, the subsequent defense-like inference (\S \ref{subsec:glm_inference}) is performed jointly over the entire claim-centered graph rather than on isolated sub-claims. This reasoning mechanism naturally reduces the impact of occasional errors in graph construction, as the model integrates information across multiple sub-claims to make robust veracity assessments. 
Finally, in the explanation summarization module, the LLM is prompted to assess the veracity of each sub-claim.  Even if the final claim prediction is incorrect, the veracity assessments of individual sub-claims still provide valuable insights, offering a fine-grained explanation that helps users better understand the underlying truthfulness of different aspects of the claim.

\begin{table}[] \small
\caption{Token usage (in millions) and estimated cost of G-Defense on LIAR-RAW and RAWFC across all LLM-driven stages. 
}
\setlength\tabcolsep{2pt}
\begin{tabular}{lcccccccccc}
\hline
\multicolumn{11}{c}{\textbf{LIAR-RAW}}           \\ \hline
\multicolumn{1}{l|}{}                & \multicolumn{2}{c|}{\textbf{train}}             & \multicolumn{2}{c|}{\textbf{eval}}               & \multicolumn{2}{c|}{\textbf{test}}              & \multicolumn{2}{c|}{\textbf{sum}}                & \multicolumn{2}{c}{\textbf{avg per claim}} \\ \hline
\multicolumn{1}{l|}{}                & in      & \multicolumn{1}{c|}{out}     & in      & \multicolumn{1}{c|}{out}     & in      & \multicolumn{1}{c|}{out}     & in       & \multicolumn{1}{c|}{out}     & in               & out             \\ \hline
\multicolumn{1}{l|}{claim decomposition}    & 0.98    & \multicolumn{1}{c|}{0.54}    & 0.12    & \multicolumn{1}{c|}{0.07}    & 0.12    & \multicolumn{1}{c|}{0.07}    & 1.22     & \multicolumn{1}{c|}{0.68}    & 0.000097         & 0.000054        \\
\multicolumn{1}{l|}{edge generation} & 2.11    & \multicolumn{1}{c|}{0.93}    & 0.27    & \multicolumn{1}{c|}{0.12}    & 0.26    & \multicolumn{1}{c|}{0.12}    & 2.64     & \multicolumn{1}{c|}{1.16}    & 0.00021          & 0.000092        \\
\multicolumn{1}{l|}{explanation generation}  & 35.33   & \multicolumn{1}{c|}{2.74}    & 4.46    & \multicolumn{1}{c|}{0.34}    & 4.42    & \multicolumn{1}{c|}{0.34}    & 44.22    & \multicolumn{1}{c|}{3.42}    & 0.003512         & 0.000272        \\
\multicolumn{1}{l|}{final explanation generation}   & -       & \multicolumn{1}{c|}{-}       & -       & \multicolumn{1}{c|}{-}       & 1.11    & \multicolumn{1}{c|}{0.38}    & 1.11     & \multicolumn{1}{c|}{0.38}    & 0.000889         & 0.000301        \\ \hdashline[1pt/1pt]
\multicolumn{1}{l|}{price}           & \$19.21 & \multicolumn{1}{c|}{\$6.32} & \$2.43 & \multicolumn{1}{c|}{\$0.80} & \$2.80 & \multicolumn{1}{c|}{\$1.37} & \$24.60 & \multicolumn{1}{c|}{\$8.46}  & \$0.0024         & \$0.0011        \\ \hline
\multicolumn{11}{c}{\textbf{RAWFC}}  \\ \hline
\multicolumn{1}{l|}{claim decomposition}    & 0.16    & \multicolumn{1}{c|}{0.09}    & 0.02    & \multicolumn{1}{c|}{0.01}    & 0.02    & \multicolumn{1}{c|}{0.01}    & 0.2      & \multicolumn{1}{c|}{0.12}    & 0.0001           & 0.000058        \\
\multicolumn{1}{l|}{edge generation} & 0.34    & \multicolumn{1}{c|}{0.15}    & 0.04    & \multicolumn{1}{c|}{0.02}    & 0.04    & \multicolumn{1}{c|}{0.02}    & 0.43     & \multicolumn{1}{c|}{0.19}    & 0.000217         & 0.000095        \\
\multicolumn{1}{l|}{explanation generation}  & 6.24    & \multicolumn{1}{c|}{0.45}    & 0.78    & \multicolumn{1}{c|}{0.06}    & 0.78    & \multicolumn{1}{c|}{0.05}    & 7.8      & \multicolumn{1}{c|}{0.56}    & 0.003924         & 0.000281        \\
\multicolumn{1}{l|}{final explanation generation}   & -        & \multicolumn{1}{c|}{-}        & -        & \multicolumn{1}{c|}{-}        & 0.18    & \multicolumn{1}{c|}{0.06}    & 0.18     & \multicolumn{1}{c|}{0.06}    & 0.0009           & 0.000301        \\ \hdashline[1pt/1pt]
\multicolumn{1}{l|}{price}           & \$3.37  & \multicolumn{1}{c|}{\$1.035} & \$0.42  & \multicolumn{1}{c|}{\$0.14} & \$0.51  & \multicolumn{1}{c|}{\$0.21}  & \$4.31   & \multicolumn{1}{c|}{\$1.40} & \$0.0026      & \$0.0011      \\ \hline
\end{tabular}
\label{tab:cost_analysis}
\end{table}

\subsection{Cost and Latency Analysis}

We estimate the API cost of G-Defense by counting the tokens consumed at each LLM-based stage. The framework relies on an LLM for four steps: claim decomposition, edge generation, competing explanation generation, and final explanation generation. In experiments, we use ``gpt-3.5-turbo-0125’’ as the default LLM backbone, priced at \$0.50 per 1M input tokens and \$1.50 per 1M output tokens\footnote{https://platform.openai.com/docs/pricing}. Table~\ref{tab:cost_analysis} reports the aggregated token usage and corresponding cost on both datasets. The total cost of running G-Defense on LIAR-RAW is approximately \$33, and the cost on RAWFC is about \$6. On average, after G-Defense is fully trained, the detection cost for a claim is approximately \$0.004. These results show that G-Defense achieves strong performance with a relatively low cost, making it feasible for practical use even at large scales. In addition, the results of G-Defense$_\text{LLaMA3.1}$ in Table~\ref{tab:veracity_results} demonstrate that G-Defense can also be effectively supported by an open-source LLM backbone, rather than relying solely on closed-source models.

In addition to API cost, the inference latency of G-Defense is mainly determined by the multi-stage API interaction, evidence retrieval, and final prediction. 
The overall latency can be approximately expressed as $T_{\text{total}} \approx T_{\text{dec}} + T_{\text{rel}} + n(T_{\text{ret}}+T_{\text{comp}}) + T_{\text{pred}} + T_{\text{final}}$, where $n$ is the number of decomposed sub-claims, and $T_{\text{dec}}$, $T_{\text{rel}}$, $T_{\text{ret}}$, $T_{\text{comp}}$, $T_{\text{pred}}$, and $T_{\text{final}}$ denote the latency of claim decomposition, relation generation, evidence retrieval, competing explanation generation, final prediction, and final explanation generation, respectively. 
Based on statistics over the RAWFC and LIAR-RAW datasets, the four API-based stages of G-Defense require around 4,000 input tokens and 430 output tokens in total for each claim on average. 
In addition, each sub-claim also requires evidence retrieval by matching candidate evidence based on cosine similarity, and the final veracity label is predicted by Flan-T5-large using all collected information. 
The end-to-end inference latency mainly grows with the number of decomposed sub-claims, since both evidence retrieval and competing explanation generation are performed at the sub-claim level. 
Nevertheless, the overall inference time remains acceptable for practical explainable veracity assessment. In this sense, G-Defense is better suited to application scenarios that emphasize prediction accuracy and explanation quality, rather than highly latency-sensitive real-time response. 
For larger-scale deployment, the latency may be further reduced by parallelizing sub-claim-level processing, caching retrieved evidence, limiting the number of decomposed sub-claims or candidate evidence, and replacing API-based steps with local models.

\vspace{6pt}
\section{Conclusion and Future Work} \label{sec:conclusion}
In this paper, we propose G-Defense, a novel graph-enhanced defense framework for explainable fake news detection with LLM. The framework first decomposes a news claim into several relevant sub-claims and then constructs a claim-centered graph to capture their dependency relationships. Then, it conducts evidence retrieval for each sub-claim and generates competing explanations for two competing veracity labels. After that, to infer the final veracity, G-Defense performs a defense-like inference over the claim-centered graph supplemented with competing explanations, comparing the quality of the competing explanations and analyzing the rich dependency relationships. Finally, an LLM is prompted with the graph and prediction result to generate a fine-grained explanation graph and a summarized textual explanation. 
Experimental results on two real-world datasets demonstrate that our proposed G-Defense achieves state-of-the-art performance in both veracity prediction and explanation quality. Extensive ablation study further verifies the effectiveness of each component in G-Defense. Human evaluation and case study further validate that the explanations generated by our framework are helpful and intuitive for identifying fake news.

In the future, we plan to broaden the applicability of G-Defense by evaluating it on other types of textual claims and potential evidence beyond datasets collected from social media platforms, such as scientific, financial, and health-related misinformation. 
This will help us further examine the framework’s robustness and generalizability across different domains and evidence sources. 
We also plan to extend G-Defense to support multi-modal fake news detection, incorporating images and videos that are increasingly common in real-world misinformation. 
Another promising direction is to explore post-trained graph-aware LLMs. Recent graph foundation models \cite{jin2024llmgraph_survey,wang2025GFM_survey,tang2024graphgpt} pretrained on large-scale graph datasets can further improve the structural reasoning of LLMs beyond the format of graph-to-text translation. These models are designed for classical graph tasks and do not directly apply to our setting, where nodes are natural-language sub-claims and edges represent logical dependencies. Nevertheless, adapting graph-aware pretraining to claim-centered graphs may further strengthen G-Defense.

\section*{Acknowledgments}
The authors would like to thank the anonymous reviewers for their valuable comments. 
This work is supported by the National Key Research and Development Program of China (No.2023YFF0905400), the National Natural Science Foundation of China (No.625B2080, No.62206233, No.U2341229 and No.62406125), the Reform Commission Foundation of Jilin Province (No.2024C003), the Fundamental and Interdisciplinary Disciplines Breakthrough Plan of the Ministry of Education of China (JYB2025XDXM903), the New Cornerstone Science Foundation through the XPLORER PRIZE, and the Guangdong Basic and Applied Basic Research Foundation (No. 2026A1515011829, No.2024A1515140144).

\bibliographystyle{ACM-Reference-Format}
\bibliography{ref}

@inproceedings{zhiwei22coling,
  author       = {Zhiwei Yang and
                  Jing Ma and
                  Hechang Chen and
                  Hongzhan Lin and
                  Ziyang Luo and
                  Yi Chang},
  editor       = {Nicoletta Calzolari and
                  Chu{-}Ren Huang and
                  Hansaem Kim and
                  James Pustejovsky and
                  Leo Wanner and
                  Key{-}Sun Choi and
                  Pum{-}Mo Ryu and
                  Hsin{-}Hsi Chen and
                  Lucia Donatelli and
                  Heng Ji and
                  Sadao Kurohashi and
                  Patrizia Paggio and
                  Nianwen Xue and
                  Seokhwan Kim and
                  Younggyun Hahm and
                  Zhong He and
                  Tony Kyungil Lee and
                  Enrico Santus and
                  Francis Bond and
                  Seung{-}Hoon Na},
  title        = {A Coarse-to-fine Cascaded Evidence-Distillation Neural Network for
                  Explainable Fake News Detection},
  booktitle    = {Proceedings of the 29th International Conference on Computational
                  Linguistics, {COLING} 2022, Gyeongju, Republic of Korea, October 12-17,
                  2022},
  pages        = {2608--2621},
  publisher    = {International Committee on Computational Linguistics},
  year         = {2022},
  url          = {https://aclanthology.org/2022.coling-1.230},
  bibsource    = {dblp computer science bibliography, https://dblp.org}
}

@article{llama,
  author       = {Hugo Touvron and
                  Thibaut Lavril and
                  Gautier Izacard and
                  Xavier Martinet and
                  Marie{-}Anne Lachaux and
                  Timoth{\'{e}}e Lacroix and
                  Baptiste Rozi{\`{e}}re and
                  Naman Goyal and
                  Eric Hambro and
                  Faisal Azhar and
                  Aur{\'{e}}lien Rodriguez and
                  Armand Joulin and
                  Edouard Grave and
                  Guillaume Lample},
  title        = {LLaMA: Open and Efficient Foundation Language Models},
  journal      = {CoRR},
  volume       = {abs/2302.13971},
  year         = {2023},
  url          = {https://doi.org/10.48550/arXiv.2302.13971},
  doi          = {10.48550/arXiv.2302.13971},
  eprinttype    = {arXiv},
  eprint       = {2302.13971},
  bibsource    = {dblp computer science bibliography, https://dblp.org}
}

@article{llama2,
  title={Llama 2: Open foundation and fine-tuned chat models},
  author={Touvron, Hugo and Martin, Louis and Stone, Kevin and Albert, Peter and Almahairi, Amjad and Babaei, Yasmine and Bashlykov, Nikolay and Batra, Soumya and Bhargava, Prajjwal and Bhosale, Shruti and others},
  journal={arXiv preprint arXiv:2307.09288},
  year={2023}
}

@inproceedings{chatgpt,
  author       = {Long Ouyang and
                  Jeffrey Wu and
                  Xu Jiang and
                  Diogo Almeida and
                  Carroll L. Wainwright and
                  Pamela Mishkin and
                  Chong Zhang and
                  Sandhini Agarwal and
                  Katarina Slama and
                  Alex Ray and
                  John Schulman and
                  Jacob Hilton and
                  Fraser Kelton and
                  Luke Miller and
                  Maddie Simens and
                  Amanda Askell and
                  Peter Welinder and
                  Paul F. Christiano and
                  Jan Leike and
                  Ryan Lowe},
  title        = {Training language models to follow instructions with human feedback},
  booktitle    = {NeurIPS},
  year         = {2022},
  url          = {http://papers.nips.cc/paper\_files/paper/2022/hash/b1efde53be364a73914f58805a001731-Abstract-Conference.html},
  bibsource    = {dblp computer science bibliography, https://dblp.org}
}

@inproceedings{dEFEND_KDD19,
  author       = {Kai Shu and
                  Limeng Cui and
                  Suhang Wang and
                  Dongwon Lee and
                  Huan Liu},
  editor       = {Ankur Teredesai and
                  Vipin Kumar and
                  Ying Li and
                  R{\'{o}}mer Rosales and
                  Evimaria Terzi and
                  George Karypis},
  title        = {dEFEND: Explainable Fake News Detection},
  booktitle    = {Proceedings of the 25th {ACM} {SIGKDD} International Conference on
                  Knowledge Discovery {\&} Data Mining, {KDD} 2019, Anchorage, AK,
                  USA, August 4-8, 2019},
  pages        = {395--405},
  publisher    = {{ACM}},
  year         = {2019},
  url          = {https://doi.org/10.1145/3292500.3330935},
  doi          = {10.1145/3292500.3330935},
  bibsource    = {dblp computer science bibliography, https://dblp.org}
}

@inproceedings{GenFE_ACL20,
  author       = {Pepa Atanasova and
                  Jakob Grue Simonsen and
                  Christina Lioma and
                  Isabelle Augenstein},
  editor       = {Dan Jurafsky and
                  Joyce Chai and
                  Natalie Schluter and
                  Joel R. Tetreault},
  title        = {Generating Fact Checking Explanations},
  booktitle    = {Proceedings of the 58th Annual Meeting of the Association for Computational
                  Linguistics, {ACL} 2020, Online, July 5-10, 2020},
  pages        = {7352--7364},
  publisher    = {Association for Computational Linguistics},
  year         = {2020},
  url          = {https://doi.org/10.18653/v1/2020.acl-main.656},
  doi          = {10.18653/v1/2020.acl-main.656},
  bibsource    = {dblp computer science bibliography, https://dblp.org}
}

@article{factllama,
  author       = {Tsun{-}Hin Cheung and
                  Kin{-}Man Lam},
  title        = {FactLLaMA: Optimizing Instruction-Following Language Models with External
                  Knowledge for Automated Fact-Checking},
  journal      = {CoRR},
  volume       = {abs/2309.00240},
  year         = {2023},
  url          = {https://doi.org/10.48550/arXiv.2309.00240},
  doi          = {10.48550/arXiv.2309.00240},
  eprinttype    = {arXiv},
  eprint       = {2309.00240},
  bibsource    = {dblp computer science bibliography, https://dblp.org}
}

@inproceedings{ma_WWW18,
  author       = {Jing Ma and
                  Wei Gao and
                  Kam{-}Fai Wong},
  editor       = {Pierre{-}Antoine Champin and
                  Fabien Gandon and
                  Mounia Lalmas and
                  Panagiotis G. Ipeirotis},
  title        = {Detect Rumor and Stance Jointly by Neural Multi-task Learning},
  booktitle    = {Companion of the The Web Conference 2018 on The Web Conference 2018,
                  {WWW} 2018, Lyon , France, April 23-27, 2018},
  pages        = {585--593},
  publisher    = {{ACM}},
  year         = {2018},
  url          = {https://doi.org/10.1145/3184558.3188729},
  doi          = {10.1145/3184558.3188729},
  bibsource    = {dblp computer science bibliography, https://dblp.org}
}

@inproceedings{EXTABS_EMNLP20_Kotonya,
  author       = {Neema Kotonya and
                  Francesca Toni},
  editor       = {Bonnie Webber and
                  Trevor Cohn and
                  Yulan He and
                  Yang Liu},
  title        = {Explainable Automated Fact-Checking for Public Health Claims},
  booktitle    = {Proceedings of the 2020 Conference on Empirical Methods in Natural
                  Language Processing, {EMNLP} 2020, Online, November 16-20, 2020},
  pages        = {7740--7754},
  publisher    = {Association for Computational Linguistics},
  year         = {2020},
  url          = {https://doi.org/10.18653/v1/2020.emnlp-main.623},
  doi          = {10.18653/v1/2020.emnlp-main.623},
  bibsource    = {dblp computer science bibliography, https://dblp.org}
}

@inproceedings{lora_ICLR22,
  author       = {Edward J. Hu and
                  Yelong Shen and
                  Phillip Wallis and
                  Zeyuan Allen{-}Zhu and
                  Yuanzhi Li and
                  Shean Wang and
                  Lu Wang and
                  Weizhu Chen},
  title        = {LoRA: Low-Rank Adaptation of Large Language Models},
  booktitle    = {The Tenth International Conference on Learning Representations, {ICLR}
                  2022, Virtual Event, April 25-29, 2022},
  publisher    = {OpenReview.net},
  year         = {2022},
  url          = {https://openreview.net/forum?id=nZeVKeeFYf9},
  bibsource    = {dblp computer science bibliography, https://dblp.org}
}

@inproceedings{DeClarE_EMNLP18,
  author       = {Kashyap Popat and
                  Subhabrata Mukherjee and
                  Andrew Yates and
                  Gerhard Weikum},
  editor       = {Ellen Riloff and
                  David Chiang and
                  Julia Hockenmaier and
                  Jun'ichi Tsujii},
  title        = {DeClarE: Debunking Fake News and False Claims using Evidence-Aware
                  Deep Learning},
  booktitle    = {Proceedings of the 2018 Conference on Empirical Methods in Natural
                  Language Processing, Brussels, Belgium, October 31 - November 4, 2018},
  pages        = {22--32},
  publisher    = {Association for Computational Linguistics},
  year         = {2018},
  url          = {https://doi.org/10.18653/v1/d18-1003},
  doi          = {10.18653/v1/d18-1003},
  bibsource    = {dblp computer science bibliography, https://dblp.org}
}

@inproceedings{conflict_wu_aaai21,
  author       = {Lianwei Wu and
                  Yuan Rao and
                  Ling Sun and
                  Wangbo He},
  title        = {Evidence Inference Networks for Interpretable Claim Verification},
  booktitle    = {Thirty-Fifth {AAAI} Conference on Artificial Intelligence, {AAAI}
                  2021, Thirty-Third Conference on Innovative Applications of Artificial
                  Intelligence, {IAAI} 2021, The Eleventh Symposium on Educational Advances
                  in Artificial Intelligence, {EAAI} 2021, Virtual Event, February 2-9,
                  2021},
  pages        = {14058--14066},
  publisher    = {{AAAI} Press},
  year         = {2021},
  url          = {https://doi.org/10.1609/aaai.v35i16.17655},
  doi          = {10.1609/aaai.v35i16.17655},
  bibsource    = {dblp computer science bibliography, https://dblp.org}
}

@inproceedings{HAN_ma_acl19,
  author       = {Jing Ma and
                  Wei Gao and
                  Shafiq R. Joty and
                  Kam{-}Fai Wong},
  editor       = {Anna Korhonen and
                  David R. Traum and
                  Llu{\'{\i}}s M{\`{a}}rquez},
  title        = {Sentence-Level Evidence Embedding for Claim Verification with Hierarchical
                  Attention Networks},
  booktitle    = {Proceedings of the 57th Conference of the Association for Computational
                  Linguistics, {ACL} 2019, Florence, Italy, July 28- August 2, 2019,
                  Volume 1: Long Papers},
  pages        = {2561--2571},
  publisher    = {Association for Computational Linguistics},
  year         = {2019},
  url          = {https://doi.org/10.18653/v1/p19-1244},
  doi          = {10.18653/v1/p19-1244},
  bibsource    = {dblp computer science bibliography, https://dblp.org}
}

@inproceedings{InstructGPT_ouyang_NIPS22,
  author       = {Long Ouyang and
                  Jeffrey Wu and
                  Xu Jiang and
                  Diogo Almeida and
                  Carroll L. Wainwright and
                  Pamela Mishkin and
                  Chong Zhang and
                  Sandhini Agarwal and
                  Katarina Slama and
                  Alex Ray and
                  John Schulman and
                  Jacob Hilton and
                  Fraser Kelton and
                  Luke Miller and
                  Maddie Simens and
                  Amanda Askell and
                  Peter Welinder and
                  Paul F. Christiano and
                  Jan Leike and
                  Ryan Lowe},
  title        = {Training language models to follow instructions with human feedback},
  booktitle    = {NeurIPS},
  year         = {2022},
  url          = {http://papers.nips.cc/paper\_files/paper/2022/hash/b1efde53be364a73914f58805a001731-Abstract-Conference.html},
  bibsource    = {dblp computer science bibliography, https://dblp.org}
}

@article{GPT-4,
  author       = {OpenAI},
  title        = {{GPT-4} Technical Report},
  journal      = {CoRR},
  volume       = {abs/2303.08774},
  year         = {2023},
  url          = {https://doi.org/10.48550/arXiv.2303.08774},
  doi          = {10.48550/arXiv.2303.08774},
  eprinttype    = {arXiv},
  eprint       = {2303.08774},
  bibsource    = {dblp computer science bibliography, https://dblp.org}
}

@article{survey_LLM_eval,
  author       = {Yupeng Chang and
                  Xu Wang and
                  Jindong Wang and
                  Yuan Wu and
                  Kaijie Zhu and
                  Hao Chen and
                  Linyi Yang and
                  Xiaoyuan Yi and
                  Cunxiang Wang and
                  Yidong Wang and
                  Wei Ye and
                  Yue Zhang and
                  Yi Chang and
                  Philip S. Yu and
                  Qiang Yang and
                  Xing Xie},
  title        = {A Survey on Evaluation of Large Language Models},
  journal      = {CoRR},
  volume       = {abs/2307.03109},
  year         = {2023},
  url          = {https://doi.org/10.48550/arXiv.2307.03109},
  doi          = {10.48550/arXiv.2307.03109},
  eprinttype    = {arXiv},
  eprint       = {2307.03109},
  bibsource    = {dblp computer science bibliography, https://dblp.org}
}

@inproceedings{exp-eval-figure_WWW23_Huang,
  author       = {Fan Huang and
                  Haewoon Kwak and
                  Jisun An},
  editor       = {Ying Ding and
                  Jie Tang and
                  Juan F. Sequeda and
                  Lora Aroyo and
                  Carlos Castillo and
                  Geert{-}Jan Houben},
  title        = {Is ChatGPT better than Human Annotators? Potential and Limitations
                  of ChatGPT in Explaining Implicit Hate Speech},
  booktitle    = {Companion Proceedings of the {ACM} Web Conference 2023, {WWW} 2023,
                  Austin, TX, USA, 30 April 2023 - 4 May 2023},
  pages        = {294--297},
  publisher    = {{ACM}},
  year         = {2023},
  url          = {https://doi.org/10.1145/3543873.3587368},
  doi          = {10.1145/3543873.3587368},
  bibsource    = {dblp computer science bibliography, https://dblp.org}
}

@inproceedings{exp-eval-metric_IJCAI23_Wang,
  author       = {Han Wang and
                  Ming Shan Hee and
                  Md. Rabiul Awal and
                  Kenny Tsu Wei Choo and
                  Roy Ka{-}Wei Lee},
  title        = {Evaluating {GPT-3} Generated Explanations for Hateful Content Moderation},
  booktitle    = {Proceedings of the Thirty-Second International Joint Conference on
                  Artificial Intelligence, {IJCAI} 2023, 19th-25th August 2023, Macao,
                  SAR, China},
  pages        = {6255--6263},
  publisher    = {ijcai.org},
  year         = {2023},
  url          = {https://doi.org/10.24963/ijcai.2023/694},
  doi          = {10.24963/ijcai.2023/694},
  bibsource    = {dblp computer science bibliography, https://dblp.org}
}

@inproceedings{LLM-eval_ACL23_Chiang,
  author       = {David Cheng{-}Han Chiang and
                  Hung{-}yi Lee},
  editor       = {Anna Rogers and
                  Jordan L. Boyd{-}Graber and
                  Naoaki Okazaki},
  title        = {Can Large Language Models Be an Alternative to Human Evaluations?},
  booktitle    = {Proceedings of the 61st Annual Meeting of the Association for Computational
                  Linguistics (Volume 1: Long Papers), {ACL} 2023, Toronto, Canada,
                  July 9-14, 2023},
  pages        = {15607--15631},
  publisher    = {Association for Computational Linguistics},
  year         = {2023},
  url          = {https://doi.org/10.18653/v1/2023.acl-long.870},
  doi          = {10.18653/v1/2023.acl-long.870},
  bibsource    = {dblp computer science bibliography, https://dblp.org}
}

@inproceedings{Nie_AAAI19,
  author       = {Yixin Nie and
                  Haonan Chen and
                  Mohit Bansal},
  title        = {Combining Fact Extraction and Verification with Neural Semantic Matching
                  Networks},
  booktitle    = {The Thirty-Third {AAAI} Conference on Artificial Intelligence, {AAAI}
                  2019, The Thirty-First Innovative Applications of Artificial Intelligence
                  Conference, {IAAI} 2019, The Ninth {AAAI} Symposium on Educational
                  Advances in Artificial Intelligence, {EAAI} 2019, Honolulu, Hawaii,
                  USA, January 27 - February 1, 2019},
  pages        = {6859--6866},
  publisher    = {{AAAI} Press},
  year         = {2019},
  url          = {https://doi.org/10.1609/aaai.v33i01.33016859},
  doi          = {10.1609/aaai.v33i01.33016859},
  bibsource    = {dblp computer science bibliography, https://dblp.org}
}

@inproceedings{ruichao_sigir22,
  author       = {Ruichao Yang and
                  Jing Ma and
                  Hongzhan Lin and
                  Wei Gao},
  editor       = {Enrique Amig{\'{o}} and
                  Pablo Castells and
                  Julio Gonzalo and
                  Ben Carterette and
                  J. Shane Culpepper and
                  Gabriella Kazai},
  title        = {A Weakly Supervised Propagation Model for Rumor Verification and Stance
                  Detection with Multiple Instance Learning},
  booktitle    = {{SIGIR} '22: The 45th International {ACM} {SIGIR} Conference on Research
                  and Development in Information Retrieval, Madrid, Spain, July 11 -
                  15, 2022},
  pages        = {1761--1772},
  publisher    = {{ACM}},
  year         = {2022},
  url          = {https://doi.org/10.1145/3477495.3531930},
  doi          = {10.1145/3477495.3531930},
  bibsource    = {dblp computer science bibliography, https://dblp.org}
}

@inproceedings{LIAR-PLUS_Alhindi_18,
  title={Where is your evidence: Improving fact-checking by justification modeling},
  author={Alhindi, Tariq and Petridis, Savvas and Muresan, Smaranda},
  booktitle={Proceedings of the first workshop on fact extraction and verification (FEVER)},
  pages={85--90},
  year={2018}
}

@inproceedings{sentence-bert_Reimers_EMNLP19,
  author       = {Nils Reimers and
                  Iryna Gurevych},
  editor       = {Kentaro Inui and
                  Jing Jiang and
                  Vincent Ng and
                  Xiaojun Wan},
  title        = {Sentence-BERT: Sentence Embeddings using Siamese BERT-Networks},
  booktitle    = {Proceedings of the 2019 Conference on Empirical Methods in Natural
                  Language Processing and the 9th International Joint Conference on
                  Natural Language Processing, {EMNLP-IJCNLP} 2019, Hong Kong, China,
                  November 3-7, 2019},
  pages        = {3980--3990},
  publisher    = {Association for Computational Linguistics},
  year         = {2019},
  url          = {https://doi.org/10.18653/v1/D19-1410},
  doi          = {10.18653/v1/D19-1410},
  bibsource    = {dblp computer science bibliography, https://dblp.org}
}

@inproceedings{hallucination_bang_arxiv23,
  author       = {Yejin Bang and
                  Samuel Cahyawijaya and
                  Nayeon Lee and
                  Wenliang Dai and
                  Dan Su and
                  Bryan Wilie and
                  Holy Lovenia and
                  Ziwei Ji and
                  Tiezheng Yu and
                  Willy Chung and
                  Quyet V. Do and
                  Yan Xu and
                  Pascale Fung},
  editor       = {Jong C. Park and
                  Yuki Arase and
                  Baotian Hu and
                  Wei Lu and
                  Derry Wijaya and
                  Ayu Purwarianti and
                  Adila Alfa Krisnadhi},
  title        = {A Multitask, Multilingual, Multimodal Evaluation of ChatGPT on Reasoning,
                  Hallucination, and Interactivity},
  booktitle    = {Proceedings of the 13th International Joint Conference on Natural
                  Language Processing and the 3rd Conference of the Asia-Pacific Chapter
                  of the Association for Computational Linguistics, {IJCNLP} 2023 -Volume
                  1: Long Papers, Nusa Dua, Bali, November 1 - 4, 2023},
  pages        = {675--718},
  publisher    = {Association for Computational Linguistics},
  year         = {2023},
  url          = {https://doi.org/10.18653/v1/2023.ijcnlp-main.45},
  doi          = {10.18653/V1/2023.IJCNLP-MAIN.45},
  bibsource    = {dblp computer science bibliography, https://dblp.org}
}

@article{RoBERTa,
  author    = {Yinhan Liu and
               Myle Ott and
               Naman Goyal and
               Jingfei Du and
               Mandar Joshi and
               Danqi Chen and
               Omer Levy and
               Mike Lewis and
               Luke Zettlemoyer and
               Veselin Stoyanov},
  title     = {RoBERTa: {A} Robustly Optimized {BERT} Pretraining Approach},
  journal   = {CoRR},
  volume    = {abs/1907.11692},
  year      = {2019},
  url       = {http://arxiv.org/abs/1907.11692},
  archivePrefix = {arXiv},
  eprint    = {1907.11692},
  bibsource = {dblp computer science bibliography, https://dblp.org}
}

@article{llm-for-eval_Chen_arxiv23,
  author       = {Yi Chen and
                  Rui Wang and
                  Haiyun Jiang and
                  Shuming Shi and
                  Ruifeng Xu},
  title        = {Exploring the Use of Large Language Models for Reference-Free Text
                  Quality Evaluation: {A} Preliminary Empirical Study},
  journal      = {CoRR},
  volume       = {abs/2304.00723},
  year         = {2023},
  url          = {https://doi.org/10.48550/arXiv.2304.00723},
  doi          = {10.48550/arXiv.2304.00723},
  eprinttype    = {arXiv},
  eprint       = {2304.00723},
  bibsource    = {dblp computer science bibliography, https://dblp.org}
}

@inproceedings{XFake_yang_www19,
  author       = {Fan Yang and
                  Shiva K. Pentyala and
                  Sina Mohseni and
                  Mengnan Du and
                  Hao Yuan and
                  Rhema Linder and
                  Eric D. Ragan and
                  Shuiwang Ji and
                  Xia (Ben) Hu},
  editor       = {Ling Liu and
                  Ryen W. White and
                  Amin Mantrach and
                  Fabrizio Silvestri and
                  Julian J. McAuley and
                  Ricardo Baeza{-}Yates and
                  Leila Zia},
  title        = {XFake: Explainable Fake News Detector with Visualizations},
  booktitle    = {The World Wide Web Conference, {WWW} 2019, San Francisco, CA, USA,
                  May 13-17, 2019},
  pages        = {3600--3604},
  publisher    = {{ACM}},
  year         = {2019},
  url          = {https://doi.org/10.1145/3308558.3314119},
  doi          = {10.1145/3308558.3314119},
  bibsource    = {dblp computer science bibliography, https://dblp.org}
}

@inproceedings{GCAN_Lu_and_Li_ACL20,
  author       = {Yi{-}Ju Lu and
                  Cheng{-}Te Li},
  editor       = {Dan Jurafsky and
                  Joyce Chai and
                  Natalie Schluter and
                  Joel R. Tetreault},
  title        = {{GCAN:} Graph-aware Co-Attention Networks for Explainable Fake News
                  Detection on Social Media},
  booktitle    = {Proceedings of the 58th Annual Meeting of the Association for Computational
                  Linguistics, {ACL} 2020, Online, July 5-10, 2020},
  pages        = {505--514},
  publisher    = {Association for Computational Linguistics},
  year         = {2020},
  url          = {https://doi.org/10.18653/v1/2020.acl-main.48},
  doi          = {10.18653/v1/2020.acl-main.48},
  bibsource    = {dblp computer science bibliography, https://dblp.org}
}

@article{LLM_biased_su_arxiv23,
  author       = {Jinyan Su and
                  Terry Yue Zhuo and
                  Jonibek Mansurov and
                  Di Wang and
                  Preslav Nakov},
  title        = {Fake News Detectors are Biased against Texts Generated by Large Language
                  Models},
  journal      = {CoRR},
  volume       = {abs/2309.08674},
  year         = {2023},
  url          = {https://doi.org/10.48550/arXiv.2309.08674},
  doi          = {10.48550/arXiv.2309.08674},
  eprinttype    = {arXiv},
  eprint       = {2309.08674},
  bibsource    = {dblp computer science bibliography, https://dblp.org}
}

@inproceedings{EFKD_survey_KotonyaT_coling20,
  author       = {Neema Kotonya and
                  Francesca Toni},
  editor       = {Donia Scott and
                  N{\'{u}}ria Bel and
                  Chengqing Zong},
  title        = {Explainable Automated Fact-Checking: {A} Survey},
  booktitle    = {Proceedings of the 28th International Conference on Computational
                  Linguistics, {COLING} 2020, Barcelona, Spain (Online), December 8-13,
                  2020},
  pages        = {5430--5443},
  publisher    = {International Committee on Computational Linguistics},
  year         = {2020},
  url          = {https://doi.org/10.18653/v1/2020.coling-main.474},
  doi          = {10.18653/v1/2020.coling-main.474},
  bibsource    = {dblp computer science bibliography, https://dblp.org}
}

@article{crowds_wisdom-Allen_2021,
  title={Scaling up fact-checking using the wisdom of crowds},
  author={Allen, Jennifer and Arechar, Antonio A and Pennycook, Gordon and Rand, David G},
  journal={Science advances},
  volume={7},
  number={36},
  pages={eabf4393},
  year={2021},
  publisher={American Association for the Advancement of Science}
}

@article{transformer,
  title={Attention is all you need},
  author={Vaswani, Ashish and Shazeer, Noam and Parmar, Niki and Uszkoreit, Jakob and Jones, Llion and Gomez, Aidan N and Kaiser, {\L}ukasz and Polosukhin, Illia},
  journal={Advances in neural information processing systems},
  volume={30},
  year={2017}
}

@inproceedings{jin_AAAI16,
  title={News verification by exploiting conflicting social viewpoints in microblogs},
  author={Jin, Zhiwei and Cao, Juan and Zhang, Yongdong and Luo, Jiebo},
  booktitle={Proceedings of the AAAI conference on artificial intelligence},
  volume={30},
  number={1},
  year={2016}
}

@inproceedings{aacl23_zhang_gao_hiss,
  author       = {Xuan Zhang and
                  Wei Gao},
  editor       = {Jong C. Park and
                  Yuki Arase and
                  Baotian Hu and
                  Wei Lu and
                  Derry Wijaya and
                  Ayu Purwarianti and
                  Adila Alfa Krisnadhi},
  title        = {Towards LLM-based Fact Verification on News Claims with a Hierarchical
                  Step-by-Step Prompting Method},
  booktitle    = {Proceedings of the 13th International Joint Conference on Natural
                  Language Processing and the 3rd Conference of the Asia-Pacific Chapter
                  of the Association for Computational Linguistics, {IJCNLP} 2023 -Volume
                  1: Long Papers, Nusa Dua, Bali, November 1 - 4, 2023},
  pages        = {996--1011},
  publisher    = {Association for Computational Linguistics},
  year         = {2023},
  url          = {https://doi.org/10.18653/v1/2023.ijcnlp-main.64},
  doi          = {10.18653/V1/2023.IJCNLP-MAIN.64},
  bibsource    = {dblp computer science bibliography, https://dblp.org}
}

@inproceedings{zhang_coling24,
  author       = {Xuan Zhang and
                  Wei Gao},
  editor       = {Nicoletta Calzolari and
                  Min{-}Yen Kan and
                  V{\'{e}}ronique Hoste and
                  Alessandro Lenci and
                  Sakriani Sakti and
                  Nianwen Xue},
  title        = {Reinforcement Retrieval Leveraging Fine-grained Feedback for Fact
                  Checking News Claims with Black-Box {LLM}},
  booktitle    = {Proceedings of the 2024 Joint International Conference on Computational
                  Linguistics, Language Resources and Evaluation, {LREC/COLING} 2024,
                  20-25 May, 2024, Torino, Italy},
  pages        = {13861--13873},
  publisher    = {{ELRA} and {ICCL}},
  year         = {2024},
  url          = {https://aclanthology.org/2024.lrec-main.1209},
  bibsource    = {dblp computer science bibliography, https://dblp.org}
}

@inproceedings{wang_www24,
  author       = {Bo Wang and
                  Jing Ma and
                  Hongzhan Lin and
                  Zhiwei Yang and
                  Ruichao Yang and
                  Yuan Tian and
                  Yi Chang},
  editor       = {Tat{-}Seng Chua and
                  Chong{-}Wah Ngo and
                  Ravi Kumar and
                  Hady W. Lauw and
                  Roy Ka{-}Wei Lee},
  title        = {Explainable Fake News Detection with Large Language Model via Defense
                  Among Competing Wisdom},
  booktitle    = {Proceedings of the {ACM} on Web Conference 2024, {WWW} 2024, Singapore,
                  May 13-17, 2024},
  pages        = {2452--2463},
  publisher    = {{ACM}},
  year         = {2024},
  url          = {https://doi.org/10.1145/3589334.3645471},
  doi          = {10.1145/3589334.3645471},
  bibsource    = {dblp computer science bibliography, https://dblp.org}
}

@article{llama3,
  title={The llama 3 herd of models},
  author={Dubey, Abhimanyu and Jauhri, Abhinav and Pandey, Abhinav and Kadian, Abhishek and Al-Dahle, Ahmad and Letman, Aiesha and Mathur, Akhil and Schelten, Alan and Yang, Amy and Fan, Angela and others},
  journal={arXiv preprint arXiv:2407.21783},
  year={2024}
}

@inproceedings{folk_emnlp23_wang_and_shu,
  author       = {Haoran Wang and
                  Kai Shu},
  editor       = {Houda Bouamor and
                  Juan Pino and
                  Kalika Bali},
  title        = {Explainable Claim Verification via Knowledge-Grounded Reasoning with
                  Large Language Models},
  booktitle    = {Findings of the Association for Computational Linguistics: {EMNLP}
                  2023, Singapore, December 6-10, 2023},
  pages        = {6288--6304},
  publisher    = {Association for Computational Linguistics},
  year         = {2023},
  url          = {https://doi.org/10.18653/v1/2023.findings-emnlp.416},
  doi          = {10.18653/V1/2023.FINDINGS-EMNLP.416},
  bibsource    = {dblp computer science bibliography, https://dblp.org}
}

@inproceedings{subquestion_emnlp22_chen,
  author       = {Jifan Chen and
                  Aniruddh Sriram and
                  Eunsol Choi and
                  Greg Durrett},
  editor       = {Yoav Goldberg and
                  Zornitsa Kozareva and
                  Yue Zhang},
  title        = {Generating Literal and Implied Subquestions to Fact-check Complex
                  Claims},
  booktitle    = {Proceedings of the 2022 Conference on Empirical Methods in Natural
                  Language Processing, {EMNLP} 2022, Abu Dhabi, United Arab Emirates,
                  December 7-11, 2022},
  pages        = {3495--3516},
  publisher    = {Association for Computational Linguistics},
  year         = {2022},
  url          = {https://doi.org/10.18653/v1/2022.emnlp-main.229},
  doi          = {10.18653/V1/2022.EMNLP-MAIN.229},
  bibsource    = {dblp computer science bibliography, https://dblp.org}
}

@inproceedings{acl24_findings_wan,
  author       = {Herun Wan and
                  Shangbin Feng and
                  Zhaoxuan Tan and
                  Heng Wang and
                  Yulia Tsvetkov and
                  Minnan Luo},
  editor       = {Lun{-}Wei Ku and
                  Andre Martins and
                  Vivek Srikumar},
  title        = {{DELL:} Generating Reactions and Explanations for LLM-Based Misinformation
                  Detection},
  booktitle    = {Findings of the Association for Computational Linguistics, {ACL} 2024,
                  Bangkok, Thailand and virtual meeting, August 11-16, 2024},
  pages        = {2637--2667},
  publisher    = {Association for Computational Linguistics},
  year         = {2024},
  url          = {https://doi.org/10.18653/v1/2024.findings-acl.155},
  doi          = {10.18653/V1/2024.FINDINGS-ACL.155},
  bibsource    = {dblp computer science bibliography, https://dblp.org}
}

@inproceedings{naacl24_yue_rag,
  author       = {Zhenrui Yue and
                  Huimin Zeng and
                  Yimeng Lu and
                  Lanyu Shang and
                  Yang Zhang and
                  Dong Wang},
  editor       = {Kevin Duh and
                  Helena G{\'{o}}mez{-}Adorno and
                  Steven Bethard},
  title        = {Evidence-Driven Retrieval Augmented Response Generation for Online
                  Misinformation},
  booktitle    = {Proceedings of the 2024 Conference of the North American Chapter of
                  the Association for Computational Linguistics: Human Language Technologies
                  (Volume 1: Long Papers), {NAACL} 2024, Mexico City, Mexico, June 16-21,
                  2024},
  pages        = {5628--5643},
  publisher    = {Association for Computational Linguistics},
  year         = {2024},
  url          = {https://doi.org/10.18653/v1/2024.naacl-long.313},
  doi          = {10.18653/V1/2024.NAACL-LONG.313},
  bibsource    = {dblp computer science bibliography, https://dblp.org}
}

@inproceedings{cikm24_nan,
  author       = {Qiong Nan and
                  Qiang Sheng and
                  Juan Cao and
                  Beizhe Hu and
                  Danding Wang and
                  Jintao Li},
  editor       = {Edoardo Serra and
                  Francesca Spezzano},
  title        = {Let Silence Speak: Enhancing Fake News Detection with Generated Comments
                  from Large Language Models},
  booktitle    = {Proceedings of the 33rd {ACM} International Conference on Information
                  and Knowledge Management, {CIKM} 2024, Boise, ID, USA, October 21-25,
                  2024},
  pages        = {1732--1742},
  publisher    = {{ACM}},
  year         = {2024},
  url          = {https://doi.org/10.1145/3627673.3679519},
  doi          = {10.1145/3627673.3679519},
  bibsource    = {dblp computer science bibliography, https://dblp.org}
}

@inproceedings{aaai24_hu,
  author       = {Beizhe Hu and
                  Qiang Sheng and
                  Juan Cao and
                  Yuhui Shi and
                  Yang Li and
                  Danding Wang and
                  Peng Qi},
  editor       = {Michael J. Wooldridge and
                  Jennifer G. Dy and
                  Sriraam Natarajan},
  title        = {Bad Actor, Good Advisor: Exploring the Role of Large Language Models
                  in Fake News Detection},
  booktitle    = {Thirty-Eighth {AAAI} Conference on Artificial Intelligence, {AAAI}
                  2024, Thirty-Sixth Conference on Innovative Applications of Artificial
                  Intelligence, {IAAI} 2024, Fourteenth Symposium on Educational Advances
                  in Artificial Intelligence, {EAAI} 2014, February 20-27, 2024, Vancouver,
                  Canada},
  pages        = {22105--22113},
  publisher    = {{AAAI} Press},
  year         = {2024},
  url          = {https://doi.org/10.1609/aaai.v38i20.30214},
  doi          = {10.1609/AAAI.V38I20.30214},
  bibsource    = {dblp computer science bibliography, https://dblp.org}
}

@inproceedings{acl2024_yue,
  author       = {Zhenrui Yue and
                  Huimin Zeng and
                  Lanyu Shang and
                  Yifan Liu and
                  Yang Zhang and
                  Dong Wang},
  editor       = {Lun{-}Wei Ku and
                  Andre Martins and
                  Vivek Srikumar},
  title        = {Retrieval Augmented Fact Verification by Synthesizing Contrastive
                  Arguments},
  booktitle    = {Proceedings of the 62nd Annual Meeting of the Association for Computational
                  Linguistics (Volume 1: Long Papers), {ACL} 2024, Bangkok, Thailand,
                  August 11-16, 2024},
  pages        = {10331--10343},
  publisher    = {Association for Computational Linguistics},
  year         = {2024},
  url          = {https://doi.org/10.18653/v1/2024.acl-long.556},
  doi          = {10.18653/V1/2024.ACL-LONG.556},
  bibsource    = {dblp computer science bibliography, https://dblp.org}
}

@inproceedings{naacl24_chen,
  author       = {Jifan Chen and
                  Grace Kim and
                  Aniruddh Sriram and
                  Greg Durrett and
                  Eunsol Choi},
  editor       = {Kevin Duh and
                  Helena G{\'{o}}mez{-}Adorno and
                  Steven Bethard},
  title        = {Complex Claim Verification with Evidence Retrieved in the Wild},
  booktitle    = {Proceedings of the 2024 Conference of the North American Chapter of
                  the Association for Computational Linguistics: Human Language Technologies
                  (Volume 1: Long Papers), {NAACL} 2024, Mexico City, Mexico, June 16-21,
                  2024},
  pages        = {3569--3587},
  publisher    = {Association for Computational Linguistics},
  year         = {2024},
  url          = {https://doi.org/10.18653/v1/2024.naacl-long.196},
  doi          = {10.18653/V1/2024.NAACL-LONG.196},
  bibsource    = {dblp computer science bibliography, https://dblp.org}
}

@article{hallucination_survey23_ji,
  author       = {Ziwei Ji and
                  Nayeon Lee and
                  Rita Frieske and
                  Tiezheng Yu and
                  Dan Su and
                  Yan Xu and
                  Etsuko Ishii and
                  Yejin Bang and
                  Andrea Madotto and
                  Pascale Fung},
  title        = {Survey of Hallucination in Natural Language Generation},
  journal      = {{ACM} Comput. Surv.},
  volume       = {55},
  number       = {12},
  pages        = {248:1--248:38},
  year         = {2023},
  url          = {https://doi.org/10.1145/3571730},
  doi          = {10.1145/3571730},
  bibsource    = {dblp computer science bibliography, https://dblp.org}
}

@article{hallucination_survey23_zhang,
  author       = {Yue Zhang and
                  Yafu Li and
                  Leyang Cui and
                  Deng Cai and
                  Lemao Liu and
                  Tingchen Fu and
                  Xinting Huang and
                  Enbo Zhao and
                  Yu Zhang and
                  Yulong Chen and
                  Longyue Wang and
                  Anh Tuan Luu and
                  Wei Bi and
                  Freda Shi and
                  Shuming Shi},
  title        = {Siren's Song in the {AI} Ocean: {A} Survey on Hallucination in Large
                  Language Models},
  journal      = {CoRR},
  volume       = {abs/2309.01219},
  year         = {2023},
  url          = {https://doi.org/10.48550/arXiv.2309.01219},
  doi          = {10.48550/ARXIV.2309.01219},
  eprinttype    = {arXiv},
  eprint       = {2309.01219},
  bibsource    = {dblp computer science bibliography, https://dblp.org}
}

@article{tacl23_ram_rag,
  author       = {Ori Ram and
                  Yoav Levine and
                  Itay Dalmedigos and
                  Dor Muhlgay and
                  Amnon Shashua and
                  Kevin Leyton{-}Brown and
                  Yoav Shoham},
  title        = {In-Context Retrieval-Augmented Language Models},
  journal      = {Trans. Assoc. Comput. Linguistics},
  volume       = {11},
  pages        = {1316--1331},
  year         = {2023},
  url          = {https://doi.org/10.1162/tacl\_a\_00605},
  doi          = {10.1162/TACL\_A\_00605},
  bibsource    = {dblp computer science bibliography, https://dblp.org}
}

@inproceedings{neurips20_lewis_rag,
  author       = {Patrick S. H. Lewis and
                  Ethan Perez and
                  Aleksandra Piktus and
                  Fabio Petroni and
                  Vladimir Karpukhin and
                  Naman Goyal and
                  Heinrich K{\"{u}}ttler and
                  Mike Lewis and
                  Wen{-}tau Yih and
                  Tim Rockt{\"{a}}schel and
                  Sebastian Riedel and
                  Douwe Kiela},
  editor       = {Hugo Larochelle and
                  Marc'Aurelio Ranzato and
                  Raia Hadsell and
                  Maria{-}Florina Balcan and
                  Hsuan{-}Tien Lin},
  title        = {Retrieval-Augmented Generation for Knowledge-Intensive {NLP} Tasks},
  booktitle    = {Advances in Neural Information Processing Systems 33: Annual Conference
                  on Neural Information Processing Systems 2020, NeurIPS 2020, December
                  6-12, 2020, virtual},
  year         = {2020},
  url          = {https://proceedings.neurips.cc/paper/2020/hash/6b493230205f780e1bc26945df7481e5-Abstract.html},
  bibsource    = {dblp computer science bibliography, https://dblp.org}
}

@article{langchain,
  title={LangChain},
  author={Harrison Chase},
  year={2022},
url={https://github.com/langchain-ai/langchain}
}

@article{rag_survey23_gao,
  author       = {Yunfan Gao and
                  Yun Xiong and
                  Xinyu Gao and
                  Kangxiang Jia and
                  Jinliu Pan and
                  Yuxi Bi and
                  Yi Dai and
                  Jiawei Sun and
                  Qianyu Guo and
                  Meng Wang and
                  Haofen Wang},
  title        = {Retrieval-Augmented Generation for Large Language Models: {A} Survey},
  journal      = {CoRR},
  volume       = {abs/2312.10997},
  year         = {2023},
  url          = {https://doi.org/10.48550/arXiv.2312.10997},
  doi          = {10.48550/ARXIV.2312.10997},
  eprinttype    = {arXiv},
  eprint       = {2312.10997},
  bibsource    = {dblp computer science bibliography, https://dblp.org}
}

@article{rag_survey24_huang,
  author       = {Yizheng Huang and
                  Jimmy Huang},
  title        = {A Survey on Retrieval-Augmented Text Generation for Large Language
                  Models},
  journal      = {CoRR},
  volume       = {abs/2404.10981},
  year         = {2024},
  url          = {https://doi.org/10.48550/arXiv.2404.10981},
  doi          = {10.48550/ARXIV.2404.10981},
  eprinttype    = {arXiv},
  eprint       = {2404.10981},
  bibsource    = {dblp computer science bibliography, https://dblp.org}
}

@article{tkde24_jin_survey_LLM-on-Graph,
  author       = {Bowen Jin and
                  Gang Liu and
                  Chi Han and
                  Meng Jiang and
                  Heng Ji and
                  Jiawei Han},
  title        = {Large Language Models on Graphs: {A} Comprehensive Survey},
  journal      = {{IEEE} Trans. Knowl. Data Eng.},
  volume       = {36},
  number       = {12},
  pages        = {8622--8642},
  year         = {2024},
  url          = {https://doi.org/10.1109/TKDE.2024.3469578},
  doi          = {10.1109/TKDE.2024.3469578},
  bibsource    = {dblp computer science bibliography, https://dblp.org}
}

@inproceedings{iclr24_fatemi_graph-encoding-for-llm,
  author       = {Bahare Fatemi and
                  Jonathan Halcrow and
                  Bryan Perozzi},
  title        = {Talk like a Graph: Encoding Graphs for Large Language Models},
  booktitle    = {The Twelfth International Conference on Learning Representations,
                  {ICLR} 2024, Vienna, Austria, May 7-11, 2024},
  publisher    = {OpenReview.net},
  year         = {2024},
  url          = {https://openreview.net/forum?id=IuXR1CCrSi},
  bibsource    = {dblp computer science bibliography, https://dblp.org}
}

@inproceedings{eacl24_findings_ye_instructGLM,
  author       = {Ruosong Ye and
                  Caiqi Zhang and
                  Runhui Wang and
                  Shuyuan Xu and
                  Yongfeng Zhang},
  editor       = {Yvette Graham and
                  Matthew Purver},
  title        = {Language is All a Graph Needs},
  booktitle    = {Findings of the Association for Computational Linguistics: {EACL}
                  2024, St. Julian's, Malta, March 17-22, 2024},
  pages        = {1955--1973},
  publisher    = {Association for Computational Linguistics},
  year         = {2024},
  url          = {https://aclanthology.org/2024.findings-eacl.132},
  bibsource    = {dblp computer science bibliography, https://dblp.org}
}

@article{Flan-T5,
  author       = {Hyung Won Chung and
                  Le Hou and
                  Shayne Longpre and
                  Barret Zoph and
                  Yi Tay and
                  William Fedus and
                  Yunxuan Li and
                  Xuezhi Wang and
                  Mostafa Dehghani and
                  Siddhartha Brahma and
                  Albert Webson and
                  Shixiang Shane Gu and
                  Zhuyun Dai and
                  Mirac Suzgun and
                  Xinyun Chen and
                  Aakanksha Chowdhery and
                  Alex Castro{-}Ros and
                  Marie Pellat and
                  Kevin Robinson and
                  Dasha Valter and
                  Sharan Narang and
                  Gaurav Mishra and
                  Adams Yu and
                  Vincent Y. Zhao and
                  Yanping Huang and
                  Andrew M. Dai and
                  Hongkun Yu and
                  Slav Petrov and
                  Ed H. Chi and
                  Jeff Dean and
                  Jacob Devlin and
                  Adam Roberts and
                  Denny Zhou and
                  Quoc V. Le and
                  Jason Wei},
  title        = {Scaling Instruction-Finetuned Language Models},
  journal      = {J. Mach. Learn. Res.},
  volume       = {25},
  pages        = {70:1--70:53},
  year         = {2024},
  url          = {https://jmlr.org/papers/v25/23-0870.html},
  bibsource    = {dblp computer science bibliography, https://dblp.org}
}

@article{arxiv23_huang_glm,
  author       = {Jin Huang and
                  Xingjian Zhang and
                  Qiaozhu Mei and
                  Jiaqi Ma},
  title        = {Can LLMs Effectively Leverage Graph Structural Information: When and
                  Why},
  journal      = {CoRR},
  volume       = {abs/2309.16595},
  year         = {2023},
  url          = {https://doi.org/10.48550/arXiv.2309.16595},
  doi          = {10.48550/ARXIV.2309.16595},
  eprinttype    = {arXiv},
  eprint       = {2309.16595},
  bibsource    = {dblp computer science bibliography, https://dblp.org}
}

@article{kdd23_chen_llm_graph,
  author       = {Zhikai Chen and
                  Haitao Mao and
                  Hang Li and
                  Wei Jin and
                  Hongzhi Wen and
                  Xiaochi Wei and
                  Shuaiqiang Wang and
                  Dawei Yin and
                  Wenqi Fan and
                  Hui Liu and
                  Jiliang Tang},
  title        = {Exploring the Potential of Large Language Models (LLMs)in Learning
                  on Graphs},
  journal      = {{SIGKDD} Explor.},
  volume       = {25},
  number       = {2},
  pages        = {42--61},
  year         = {2023},
  url          = {https://doi.org/10.1145/3655103.3655110},
  doi          = {10.1145/3655103.3655110},
  bibsource    = {dblp computer science bibliography, https://dblp.org}
}

@article{llm_survey_zhao_23,
  title={A survey of large language models},
  author={Zhao, Wayne Xin and Zhou, Kun and Li, Junyi and Tang, Tianyi and Wang, Xiaolei and Hou, Yupeng and Min, Yingqian and Zhang, Beichen and Zhang, Junjie and Dong, Zican and others},
  journal={arXiv preprint arXiv:2303.18223},
  year={2023}
}

@inproceedings{RAG_survey_Fan_KDD24,
  title={A survey on rag meeting llms: Towards retrieval-augmented large language models},
  author={Fan, Wenqi and Ding, Yujuan and Ning, Liangbo and Wang, Shijie and Li, Hengyun and Yin, Dawei and Chua, Tat-Seng and Li, Qing},
  booktitle={Proceedings of the 30th ACM SIGKDD Conference on Knowledge Discovery and Data Mining},
  pages={6491--6501},
  year={2024}
}

@inproceedings{RAG_recom_wu_kdd24,
  author       = {Junda Wu and
                  Cheng{-}Chun Chang and
                  Tong Yu and
                  Zhankui He and
                  Jianing Wang and
                  Yupeng Hou and
                  Julian J. McAuley},
  editor       = {Ricardo Baeza{-}Yates and
                  Francesco Bonchi},
  title        = {CoRAL: Collaborative Retrieval-Augmented Large Language Models Improve
                  Long-tail Recommendation},
  booktitle    = {Proceedings of the 30th {ACM} {SIGKDD} Conference on Knowledge Discovery
                  and Data Mining, {KDD} 2024, Barcelona, Spain, August 25-29, 2024},
  pages        = {3391--3401},
  publisher    = {{ACM}},
  year         = {2024},
  url          = {https://doi.org/10.1145/3637528.3671901},
  doi          = {10.1145/3637528.3671901},
  bibsource    = {dblp computer science bibliography, https://dblp.org}
}

@inproceedings{RAG_recom_Lu_ACL21,
  author       = {Yu Lu and
                  Junwei Bao and
                  Yan Song and
                  Zichen Ma and
                  Shuguang Cui and
                  Youzheng Wu and
                  Xiaodong He},
  editor       = {Chengqing Zong and
                  Fei Xia and
                  Wenjie Li and
                  Roberto Navigli},
  title        = {RevCore: Review-Augmented Conversational Recommendation},
  booktitle    = {Findings of the Association for Computational Linguistics: {ACL/IJCNLP}
                  2021, Online Event, August 1-6, 2021},
  series       = {Findings of {ACL}},
  volume       = {{ACL/IJCNLP} 2021},
  pages        = {1161--1173},
  publisher    = {Association for Computational Linguistics},
  year         = {2021},
  url          = {https://doi.org/10.18653/v1/2021.findings-acl.99},
  doi          = {10.18653/V1/2021.FINDINGS-ACL.99},
  bibsource    = {dblp computer science bibliography, https://dblp.org}
}

@inproceedings{rag_QA_Guu_ICML20,
  author       = {Kelvin Guu and
                  Kenton Lee and
                  Zora Tung and
                  Panupong Pasupat and
                  Ming{-}Wei Chang},
  title        = {Retrieval Augmented Language Model Pre-Training},
  booktitle    = {Proceedings of the 37th International Conference on Machine Learning,
                  {ICML} 2020, 13-18 July 2020, Virtual Event},
  series       = {Proceedings of Machine Learning Research},
  volume       = {119},
  pages        = {3929--3938},
  publisher    = {{PMLR}},
  year         = {2020},
  url          = {http://proceedings.mlr.press/v119/guu20a.html},
  bibsource    = {dblp computer science bibliography, https://dblp.org}
}

@article{rag_fact_checking_Izacard_23,
  author       = {Gautier Izacard and
                  Patrick S. H. Lewis and
                  Maria Lomeli and
                  Lucas Hosseini and
                  Fabio Petroni and
                  Timo Schick and
                  Jane Dwivedi{-}Yu and
                  Armand Joulin and
                  Sebastian Riedel and
                  Edouard Grave},
  title        = {Atlas: Few-shot Learning with Retrieval Augmented Language Models},
  journal      = {J. Mach. Learn. Res.},
  volume       = {24},
  pages        = {251:1--251:43},
  year         = {2023},
  url          = {https://jmlr.org/papers/v24/23-0037.html},
  bibsource    = {dblp computer science bibliography, https://dblp.org}
}

@article{zeng_NAACL24_rag,
  author       = {Shenglai Zeng and
                  Jiankun Zhang and
                  Bingheng Li and
                  Yuping Lin and
                  Tianqi Zheng and
                  Dante Everaert and
                  Hanqing Lu and
                  Hui Liu and
                  Yue Xing and
                  Monica Xiao Cheng and
                  Jiliang Tang},
  title        = {Towards Knowledge Checking in Retrieval-augmented Generation: {A}
                  Representation Perspective},
  journal      = {CoRR},
  volume       = {abs/2411.14572},
  year         = {2024},
  url          = {https://doi.org/10.48550/arXiv.2411.14572},
  doi          = {10.48550/ARXIV.2411.14572},
  eprinttype    = {arXiv},
  eprint       = {2411.14572},
  bibsource    = {dblp computer science bibliography, https://dblp.org}
}

@article{distilbert,
  author       = {Victor Sanh and
                  Lysandre Debut and
                  Julien Chaumond and
                  Thomas Wolf},
  title        = {DistilBERT, a distilled version of {BERT:} smaller, faster, cheaper
                  and lighter},
  journal      = {CoRR},
  volume       = {abs/1910.01108},
  year         = {2019},
  url          = {http://arxiv.org/abs/1910.01108},
  eprinttype    = {arXiv},
  eprint       = {1910.01108},
  bibsource    = {dblp computer science bibliography, https://dblp.org}
}

@article{guo2025deepseek_r1,
  title={Deepseek-r1: Incentivizing reasoning capability in llms via reinforcement learning},
  author={Guo, Daya and Yang, Dejian and Zhang, Haowei and Song, Junxiao and Zhang, Ruoyu and Xu, Runxin and Zhu, Qihao and Ma, Shirong and Wang, Peiyi and Bi, Xiao and others},
  journal={arXiv preprint arXiv:2501.12948},
  year={2025}
}

@misc{hu_naacl25_analysis_claimdecomposition,
      title={Decomposition Dilemmas: Does Claim Decomposition Boost or Burden Fact-Checking Performance?}, 
      author={Qisheng Hu and Quanyu Long and Wenya Wang},
      year={2024},
      eprint={2411.02400},
      archivePrefix={arXiv},
      primaryClass={cs.IR},
      url={https://arxiv.org/abs/2411.02400}, 
}

@article{stance_rumor_tois24,
  author       = {Jun Li and
                  Yi Bin and
                  Yunshan Ma and
                  Yang Yang and
                  Zi Huang and
                  Tat{-}Seng Chua},
  title        = {Filter-based Stance Network for Rumor Verification},
  journal      = {{ACM} Trans. Inf. Syst.},
  volume       = {42},
  number       = {4},
  pages        = {108:1--108:28},
  year         = {2024},
  url          = {https://doi.org/10.1145/3649462},
  doi          = {10.1145/3649462},
  bibsource    = {dblp computer science bibliography, https://dblp.org}
}

@article{FND_tois,
  author       = {Yang Liu and
                  Yi{-}fang Brook Wu},
  title        = {{FNED:} {A} Deep Network for Fake News Early Detection on Social Media},
  journal      = {{ACM} Trans. Inf. Syst.},
  volume       = {38},
  number       = {3},
  pages        = {25:1--25:33},
  year         = {2020},
  url          = {https://doi.org/10.1145/3386253},
  doi          = {10.1145/3386253},
  bibsource    = {dblp computer science bibliography, https://dblp.org}
}

@article{RAG_QA_tois,
  title={Fit-rag: Black-box rag with factual information and token reduction},
  author={Mao, Yuren and Dong, Xuemei and Xu, Wenyi and Gao, Yunjun and Wei, Bin and Zhang, Ying},
  journal={ACM Transactions on Information Systems},
  volume={43},
  number={2},
  pages={1--27},
  year={2025},
  url={https://doi.org/10.1145/367695},
  publisher={ACM New York, NY}
}

@article{EFND_TOIS,
  title={End-to-End Explainable Fake News Detection Via Evidence-Claim Variational Causal Inference},
  author={Wang, Jinguang and Qian, Shengsheng and Hu, Jun and Dong, Wenxiang and Huang, Xudong and Hong, Richang},
  journal={ACM Transactions on Information Systems},
  year={2025},
  url={https://doi.org/10.1145/3728462},
  publisher={ACM New York, NY}
}

@inproceedings{muser,
  author       = {Hao Liao and
                  Jiahao Peng and
                  Zhanyi Huang and
                  Wei Zhang and
                  Guanghua Li and
                  Kai Shu and
                  Xing Xie},
  editor       = {Ambuj K. Singh and
                  Yizhou Sun and
                  Leman Akoglu and
                  Dimitrios Gunopulos and
                  Xifeng Yan and
                  Ravi Kumar and
                  Fatma Ozcan and
                  Jieping Ye},
  title        = {{MUSER:} {A} MUlti-Step Evidence Retrieval Enhancement Framework for
                  Fake News Detection},
  booktitle    = {Proceedings of the 29th {ACM} {SIGKDD} Conference on Knowledge Discovery
                  and Data Mining, {KDD} 2023, Long Beach, CA, USA, August 6-10, 2023},
  pages        = {4461--4472},
  publisher    = {{ACM}},
  year         = {2023},
  url          = {https://doi.org/10.1145/3580305.3599873},
  doi          = {10.1145/3580305.3599873},
  bibsource    = {dblp computer science bibliography, https://dblp.org}
}

@inproceedings{castillo2011information,
  title={Information credibility on twitter},
  author={Castillo, Carlos and Mendoza, Marcelo and Poblete, Barbara},
  booktitle={Proceedings of the 20th international conference on World wide web},
  pages={675--684},
  year={2011}
}

@article{huang2025survey_hallucination,
  title={A survey on hallucination in large language models: Principles, taxonomy, challenges, and open questions},
  author={Huang, Lei and Yu, Weijiang and Ma, Weitao and Zhong, Weihong and Feng, Zhangyin and Wang, Haotian and Chen, Qianglong and Peng, Weihua and Feng, Xiaocheng and Qin, Bing and others},
  journal={ACM Transactions on Information Systems},
  volume={43},
  number={2},
  pages={1--55},
  year={2025},
  publisher={ACM New York, NY}
}

@article{zhao2023survey_llm,
  title={A survey of large language models},
  author={Zhao, Wayne Xin and Zhou, Kun and Li, Junyi and Tang, Tianyi and Wang, Xiaolei and Hou, Yupeng and Min, Yingqian and Zhang, Beichen and Zhang, Junjie and Dong, Zican and others},
  journal={arXiv preprint arXiv:2303.18223},
  volume={1},
  number={2},
  year={2023}
}

@article{rishi2022survey_foundation,
  author       = {Rishi Bommasani and
                  Drew A. Hudson and
                  Ehsan Adeli and
                  Russ B. Altman and
                  Simran Arora and
                  Sydney von Arx and
                  Michael S. Bernstein and
                  Jeannette Bohg and
                  Antoine Bosselut and
                  Emma Brunskill and
                  Erik Brynjolfsson and
                  Shyamal Buch and
                  Dallas Card and
                  Rodrigo Castellon and
                  Niladri S. Chatterji and
                  Annie S. Chen and
                  Kathleen Creel and
                  Jared Quincy Davis and
                  Dorottya Demszky and
                  Chris Donahue and
                  Moussa Doumbouya and
                  Esin Durmus and
                  Stefano Ermon and
                  John Etchemendy and
                  Kawin Ethayarajh and
                  Li Fei{-}Fei and
                  Chelsea Finn and
                  Trevor Gale and
                  Lauren E. Gillespie and
                  Karan Goel and
                  Noah D. Goodman and
                  Shelby Grossman and
                  Neel Guha and
                  Tatsunori Hashimoto and
                  Peter Henderson and
                  John Hewitt and
                  Daniel E. Ho and
                  Jenny Hong and
                  Kyle Hsu and
                  Jing Huang and
                  Thomas Icard and
                  Saahil Jain and
                  Dan Jurafsky and
                  Pratyusha Kalluri and
                  Siddharth Karamcheti and
                  Geoff Keeling and
                  Fereshte Khani and
                  Omar Khattab and
                  Pang Wei Koh and
                  Mark S. Krass and
                  Ranjay Krishna and
                  Rohith Kuditipudi and
                  et al.},
  title        = {On the Opportunities and Risks of Foundation Models},
  journal      = {CoRR},
  volume       = {abs/2108.07258},
  year         = {2021},
  url          = {https://arxiv.org/abs/2108.07258},
  eprinttype    = {arXiv},
  eprint       = {2108.07258},
  bibsource    = {dblp computer science bibliography, https://dblp.org}
}

@inproceedings{tang2024graphgpt,
  title={Graphgpt: Graph instruction tuning for large language models},
  author={Tang, Jiabin and Yang, Yuhao and Wei, Wei and Shi, Lei and Su, Lixin and Cheng, Suqi and Yin, Dawei and Huang, Chao},
  booktitle={Proceedings of the 47th International ACM SIGIR Conference on Research and Development in Information Retrieval},
  pages={491--500},
  year={2024}
}

@article{wang2025GFM_survey,
  title={Graph Foundation Models: A Comprehensive Survey},
  author={Wang, Zehong and Liu, Zheyuan and Ma, Tianyi and Li, Jiazheng and Zhang, Zheyuan and Fu, Xingbo and Li, Yiyang and Yuan, Zhengqing and Song, Wei and Ma, Yijun and others},
  journal={arXiv preprint arXiv:2505.15116},
  year={2025}
}

@article{jin2024llmgraph_survey,
  title={Large language models on graphs: A comprehensive survey},
  author={Jin, Bowen and Liu, Gang and Han, Chi and Jiang, Meng and Ji, Heng and Han, Jiawei},
  journal={IEEE Transactions on Knowledge and Data Engineering},
  year={2024},
  publisher={IEEE}
}

@inproceedings{humaneval_IAA,
  title={Rethinking the agreement in human evaluation tasks},
  author={Amidei, Jacopo and Piwek, Paul and Willis, Alistair},
  booktitle={Proceedings of the 27th International Conference on Computational Linguistics},
  pages={3318--3329},
  year={2018}
}

@inproceedings{stanojevic2014beer,
  title={Beer: Better evaluation as ranking},
  author={Stanojevi{\'c}, Milo{\v{s}} and Sima’an, Khalil},
  booktitle={Proceedings of the Ninth Workshop on Statistical Machine Translation},
  pages={414--419},
  year={2014}
}

@inproceedings{riley2024finding,
  title={Finding replicable human evaluations via stable ranking probability},
  author={Riley, Parker and Deutsch, Daniel and Foster, George and Ratnakar, Viresh and Dabirmoghaddam, Ali and Freitag, Markus},
  booktitle={Proceedings of the 2024 Conference of the North American Chapter of the Association for Computational Linguistics: Human Language Technologies (Volume 1: Long Papers)},
  pages={4908--4919},
  year={2024}
}

\appendix

\section{Prompt Designing} \label{app:prompt}
The prompts in our framework follow a simple and task-oriented design principle. Specifically, they are mainly used to specify the task objective, the relevant information to consider, and the desired output format. We intentionally avoid heavy prompt engineering or elaborate prompt tuning to keep the framework simple and broadly applicable. 
The details of prompts introduced in \S \ref{sec:approach} and \S \ref{sec:exp} are listed below.

\begin{tcolorbox}[title={Prompt used for claim decomposition, $I^{v}$ (in \S \ref{subsec:graph_construction})},
colback=gray!10, coltitle=black, colframe=gray!30, boxrule=0.5mm, arc=2mm, left=2mm, right=2mm]
You are a fake news detection assistant. Your primary function is that, given a complex news claim, based on its content, please break it down into several sub-claims relevant to the veracity of the claim. Note that directly generate sub-claims in short and clear manner, avoiding redundancy. 

\# News claim: $c$
\end{tcolorbox}

\begin{tcolorbox}[title={Prompt used for edge generation, $I^{e}$ (in \S \ref{subsec:graph_construction})},
colback=gray!10, coltitle=black, colframe=gray!30, boxrule=0.5mm, arc=2mm, left=2mm, right=2mm]
You are a graph construction assistant. Your primary function is that, given a complex news claim (index 0), and several sub-claims (index starting from 1) decomposed from the news claim, please use logical relationships to construct a reasoning graph that reflects how the sub-claims contribute to the truth of the overall claim. \\
Note: Your output should be in a dictionary format with the keys: ("analysis", "edges"). 
You should generate directed edges using the indexes of the claims, formatted as a list like [(1, 0), (2, 3), (3, 0)]. \\
\# News claim: 0. $c$ \\
\# Sub-claims: 1. $c_1$; ...; n. $c_n$.
\end{tcolorbox}

\begin{tcolorbox}[title={Prompt used for final explanation summarization, $I^{s}$, (in \S \ref{subsec:explanation_sum})},
coltitle=black, colback=gray!10, colframe=gray!30, boxrule=0.5mm, arc=2mm, left=2mm, right=2mm]
Given a claim, its corresponding claim-centered graph, and a veracity label $y^*$, please give me the veracity prediction about each sub-claim, and a streamlined rationale associate with the claim for how it is reasoned as $y^*$. \\
\# Claim-centered graph: $I^a(\rmH_{\tilde \gG}, \tilde \gV)$\\
Your response should be a Python dictionary with the following structure: \\
\{\\
\mbox{\hspace{1em}}"sub-claims-veracity": \\
\mbox{\hspace{1em}}\{\\
\mbox{\hspace{2em}}"sub-claim 1": \{\\
\mbox{\hspace{3em}}"reasoning": "Your analysis about the veracity of sub-claim 1",\\
\mbox{\hspace{3em}}"prediction": "True/False"\\
\mbox{\hspace{2em}}\},\\
\mbox{\hspace{2em}}...,\\
\mbox{\hspace{2em}}"sub-claim n": \{\\
\mbox{\hspace{3em}}"reasoning": "Your analysis about the veracity of sub-claim n.",\\
\mbox{\hspace{3em}}"prediction": "True/False"\\
\mbox{\hspace{2em}}\}\\
\mbox{\hspace{1em}}\},\\
\mbox{\hspace{1em}}"final-explanation": ""\\
\}
\end{tcolorbox}

\begin{tcolorbox}[title={Prompt used for background generation in G-Defense{$_\text{background}$},~ $I^{b}$, (in \S \ref{subsec:eval_veracity})},
coltitle=black, colback=gray!10, colframe=gray!30, boxrule=0.5mm, arc=2mm, left=2mm, right=2mm]
You have been specially designed to perform objective contextual analysis for the fake news detection task.
Your primary function is that, according to a news claim and some sentences related to it, please provide a streamlined contextual analysis that helps understand the background, circumstances, or perspectives related to the claim.
Your goal is to summarize the key contextual information — such as background facts, timelines, participants, related events, or uncertainties —that could help people interpret or understand the claim, without expressing any stance on its truthfulness.
Note: Do not repeat the claim itself, and do not imply or indicate any truth judgment.
Just directly output the short and clear contextual rationale. \\
Given a claim $c_i$ and a set of retrieved relevant reports {$s_1, s_2, ...$},
please provide a short and clear objective contextual analysis that presents factual background or explanatory information helping to understand the claim, without expressing any stance on its veracity.
\end{tcolorbox}

\begin{tcolorbox}[title={Prompt used for hyperedge generation in G-Defense{$_\text{hyper}$}, ~$I^{h}$ (in \S \ref{subsec:eval_veracity})},
colback=gray!10, coltitle=black, colframe=gray!30, boxrule=0.5mm, arc=2mm, left=2mm, right=2mm]
You are a hypergraph construction assistant.
Your primary function is that, given a complex news claim (index 0), and several sub-claims (index starting from 1) decomposed from the news claim, 
please use topic relationship to construct a reasoning graph that reflects how the sub-claims contribute to the truth of the overall claim. 
A hyperedge represents a higher-level topic or reasoning unit that connects multiple related claims (e.g., several sub-claims that jointly address one aspect of the main claim). 
Unlike simple pairwise edges, each hyperedge can include more than two nodes to capture shared themes or collective reasoning.\\
Note: Your output should be in a dictionary format with the keys: ("analysis", "hyperedges").
You should generate hyperedges using the indexes of the claims, formatted as a list of lists like [[1, 2, 0], [3, 4, 5, 0], [5, 6, 0]].\\
\# News claim: 0. $c$ \\
\# Sub-claims: 1. $c_1$; ...; n. $c_n$.
\end{tcolorbox}

\begin{tcolorbox}[title={Prompt used for enhanced claim decomposition in G-Defense{$_\text{decomp+}$}, $I^{v+}$ (in \S \ref{subsec:eval_veracity})},
colback=gray!10, coltitle=black, colframe=gray!30, boxrule=0.5mm, arc=2mm, left=2mm, right=2mm]
You are a fake news detection assistant.
Your primary function is to analyze a complex news claim and decompose it into several clear, atomic sub-claims that collectively capture the full meaning and verifiability of the claim.
1. Decomposition Strategies\\
When analyzing the claim, adopt systematic reasoning strategies to ensure a complete and logically structured breakdown:\\
- Logical decomposition: Separate the claim into distinct factual assertions that can be verified independently.\\
- Causal reasoning: Identify cause–effect or condition–result relations.\\
- Hierarchical reasoning: Distinguish between general and specific statements or between main ideas and supporting facts.\\
- Factual reasoning: Focus on objectively testable propositions rather than opinions or implications.\\
2. News and Communication Cues\\
Incorporate journalistic awareness to better capture how information is presented in news texts:\\
- Recognize news structures such as headlines, reported statements, and factual details.\\
- Distinguish reported speech or attribution (e.g., “X said that...”) from factual assertions about reality.\\
- Identify quantitative or comparative claims (e.g., numbers, rates, trends) and evaluative expressions (e.g., moral or emotional framing).\\
- If the claim includes a source or attribution, include it as part of the sub-claim (e.g., “Officials claimed that...” or “According to reports,...”).\\
- When the claim involves ambiguous or questionable sources, make that explicit (e.g., “It is claimed, without evidence, that...”).\\
3. Dependency Cues\\
If there are logical dependencies among sub-claims, make them explicit:\\
- Use connectors such as “if,” “because,” “therefore,” or “as a result.”\\
- Reflect the logical flow between causes, conditions, and consequences.\\
4. Factual Dimensions\\
Ensure coverage of the key fact-checking dimensions as appropriate — such as who, what, when, where, why/how, and consequences. \\
5. Output Requirements\\
Each sub-claim should be concise (1–2 sentences), self-contained, and verifiable.\\
Avoid redundancy and overlapping information.\\
Base all sub-claims strictly on the information stated or clearly implied in the original claim; do not introduce new facts or assumptions.\\
Output only the numbered sub-claims list, without any additional explanations or commentary. \\
\# News claim: $c$
\end{tcolorbox}

\section{Validation of Assumption} \label{app:validation}

This section aims to validate our assumption that the side aligned with the gold veracity label usually provides evidence with higher informativeness and soundness, which in turn enables the LLM to generate a stronger explanation than the competing side. To this end, we compare the quality of competing explanations under different explanation-generation settings, including different backbone LLMs and prompt formulations. 
As shown in Table 7, the explanations conditioned on the gold-consistent label generally achieve higher scores in informativeness and soundness than those conditioned on the competing label. Specifically, for false claims, the false-oriented explanations are usually rated more favorably than the true-oriented ones; for true claims, the true-oriented explanations consistently perform better; and for half-true claims, the two sides are relatively closer. These results provide quantitative evidence supporting our assumption. 
Moreover, although the magnitude of the gap varies across settings, the same overall tendency can still be observed across different backbone LLMs and when replacing the original reasoning prompt $I^r$ with its simplified version $\tilde{I^r}$, which provides further evidence for our assumption.

\begin{tcolorbox}[title={Simplified prompt used for veracity-oriented explanation generation, $\tilde{I^r}$},
coltitle=black, colback=gray!10, colframe=gray!30, boxrule=0.5mm, arc=2mm, left=2mm, right=2mm]
Given a claim: $c_i$, a veracity label $\tilde y$, please give me a streamlined rationale associated with the claim for how it is reasoned as $\tilde y$. Below are some sentences that may be helpful for the rationale, but they are mixed with noise: $E_i$. 
\end{tcolorbox}

\begin{table}[t]
\centering
\caption{Explanations evaluation results of competing explanations using a 5-Point Likert scale rating by GPT-4 on RAWFC's test set. LLM denotes the backbone model used to generate competing explanations, and Prompt denotes the corresponding explanation-generation prompt. False/Half/True indicate the gold veracity label of the claim, while F/ T under each group indicate whether the generated explanation is conditioned on the false or true prior label. }
\setlength{\tabcolsep}{4pt}
\begin{tabular}{cccccccccc}
\toprule
\multirow{2}{*}{LLM} & \multirow{2}{*}{Prompt} & \multirow{2}{*}{Metric}
& \multicolumn{2}{c}{False}
& \multicolumn{2}{c}{Half}
& \multicolumn{2}{c}{True} \\
\cmidrule(lr){4-5}\cmidrule(lr){6-7}\cmidrule(lr){8-9}
& & & F & T & F & T & F & T \\
\midrule
\multirow{2}{*}{GPT-3.5} & \multirow{2}{*}{$I^{r}$}
& Informativeness & 3.97 & 3.06 & 3.87 & 3.62 & 3.75 & 3.96 \\
& & Soundness     & 4.86 & 2.83 & 3.61 & 3.45 & 4.30 & 4.60 \\
\midrule
\multirow{2}{*}{LLaMA3.1} & \multirow{2}{*}{$I^{r}$}
& Informativeness & 4.03 & 3.36 & 3.75 & 3.73 & 3.57 & 4.15 \\
& & Soundness & 4.45 & 2.95 & 4.01 & 3.90 & 3.57 & 4.33 \\
\midrule
\multirow{2}{*}{GPT-3.5} & \multirow{2}{*}{$\tilde{I^{r}}$}
& Informativeness & 4.23 & 3.50 & 4.04 & 3.88 & 4.04 & 4.36 \\
& & Soundness & 4.78 & 3.12 & 3.99 & 3.82 & 4.34 & 4.84 \\
\bottomrule
\end{tabular}
\label{tab:assumption_eval}
\end{table}

\section{Human Evaluation Reliability Analysis} \label{app:humaneval_reliability}
To further assess the reliability of the human evaluation, we conduct additional statistical analyses based on the annotations from five annotators over 30 claims. Specifically, we report both Krippendorff’s $\alpha$ and Kendall’s $W$ for the four evaluation dimensions introduced in \S \ref{subsec:eval_explanation}: misleadingness, informativeness, soundness, and readability.
For Krippendorff’s $\alpha$, the raw values are 0.36, 0.10, 0.13, and 0.12, respectively. Since explanation evaluation is inherently subjective, annotators may use different but internally consistent scoring scales, which can reduce agreement on absolute ratings. This is consistent with prior findings that annotator disagreement is common in subjective nature language generation evaluation and does not necessarily invalidate the evaluation itself \cite{humaneval_IAA}. To further examine this effect, we also compute normalized $\alpha$ values, which are 0.37, 0.24, 0.30, and 0.23, respectively. The increase after normalization suggests that part of the disagreement comes from differences in rating scale usage rather than complete disagreement in judgment.
Because Krippendorff’s $\alpha$ mainly reflects agreement on absolute scores, it may underestimate consistency in the relative ranking of methods \cite{stanojevic2014beer,riley2024finding}. We therefore additionally compute Kendall’s $W$, which measures agreement on method rankings across annotators. The resulting Kendall’s $W$ values are 0.79 for misleadingness, 0.66 for informativeness, 0.69 for soundness, and 0.64 for readability, indicating moderate to strong ranking agreement. Under the chi-square approximation, all four values are statistically significant ($p<0.05$).
Overall, these two metrics provide complementary views of reliability. Krippendorff’s $\alpha$ measures agreement on absolute ratings, while Kendall’s $W$ measures agreement on relative rankings. Taken together, the results suggest that although absolute-score agreement is limited by the subjective nature of explanation evaluation, the relative ordering of methods is reasonably stable across annotators, supporting the reliability of our human evaluation.

\end{document}